# A graph representation of molecular ensembles for polymer property prediction

Matteo Aldeghi [1], Connor W. Coley [1,2,*]

[1] Department of Chemical Engineering, Massachusetts Institute of Technology, Cambridge, MA 02139, USA

[2] Department of Electrical Engineering and Computer Science, Massachusetts Institute of Technology, Cambridge, MA 02139, USA

[*] Email: ccoley@mit.edu

Synthetic polymers are versatile and widely used materials. Similar to small organic molecules, a large chemical space of such materials is hypothetically accessible. Computational property prediction and virtual screening can accelerate polymer design by prioritizing candidates expected to have favorable properties. However, in contrast to organic molecules, polymers are often not well-defined single structures but an ensemble of similar molecules, which poses unique challenges to traditional chemical representations and machine learning approaches. Here, we introduce a graph representation of molecular ensembles and an associated graph neural network architecture that is tailored to polymer property prediction. We demonstrate that this approach captures critical features of polymeric materials, like chain architecture, monomer stoichiometry, and degree of polymerization, and achieves superior accuracy to off-the-shelf cheminformatics methodologies. While doing so, we built a dataset of simulated electron affinity and ionization potential values for >40k polymers with varying monomer composition, stoichiometry, and chain architecture, which may be used in the development of other tailored machine learning approaches. The dataset and machine learning models presented in this work pave the path toward new classes of algorithms for polymer informatics and, more broadly, introduce a framework for the modeling of molecular ensembles.

## Introduction

Synthetic polymers are key components of numerous commodities and play an essential role in our daily lives, with applications ranging from clothing, to electronics and construction, and are used in industries as diverse as automotive, energy, and healthcare.[1–10] This versatility is due to the wide range of properties achievable by tuning a polymer's chemical composition and architecture. The identification of novel copolymers for the delivery of therapeutics cargos[11–19], or for energy harvesting and storage[6,20–24], are examples of active areas of research that rely on the availability of a broad range of polymer chemistries.

Machine learning (ML) is now playing a significant role in supporting the discovery and synthesis of new functional organic molecules with specialized applications, thanks to its ability to capture subtle chemical patterns when enough data is available. The field of polymer informatics has also attracted increasing attention, with a number of studies demonstrating the use of ML for the prediction of thermal[25–32], thermodynamic[25,33–35], electronic[36–41], optical[38,42,43], and mechanical[38,44] properties of polymers and copolymers. However, while many specialized machine learning approaches have been developed for molecules and sequence-defined polymers like proteins and peptides, polymers characterized by molecular ensembles still rely on off-the-shelf cheminformatics approaches designed for single molecules. This work focuses specifically on the latter class of materials, which cover a considerable fraction of synthetic and natural polymers.

A major challenge in the development of bespoke ML models for polymer property prediction is the lack of a general polymer representation[45–49]. In fact, almost all ML models currently used for polymer property predictions do not capture the ensemble nature of the polymeric material. The vast majority of past studies have relied on molecular representations of repeating units alone, even though such approaches cannot distinguish between alternating, random, block, or graft copolymers. Recent work has tried to obviate this issue by creating cyclic oligomers from which structural fingerprints can be derived.[50] However, this approach would still struggle to distinguish different chain architectures or capture the ensemble of possible monomer sequences.

The challenge of identifying a general polymer representation stems from the fact that, contrary to small organic molecules, many polymers are stochastic objects whose properties emerge from the ensemble of molecules that they comprise. A representation that captures this ensemble nature is thus needed to develop tailored and broadly applicable ML models for polymer property prediction. Recently, text-based representations that try to capture this unique aspect of polymer chemistry have been developed. BigSMILES is a text-based representation that builds upon the simplified molecular-input line-entry system (SMILES) representation and is designed specifically to describe the stochastic nature of polymer molecules.[51] Yet, language models based on text-based representations are data inefficient, such that they generally require extensive pretraining, data augmentation, or extremely large dataset sizes to be successful in cheminformatics[52,53]. Representations that more directly capture chemical structure, like fingerprints and graphs, are thus preferred for learning tasks as they typically outperform language models in property prediction tasks when provided with the same amount of data.

In this work, we report the development and validation of a graph-based representation of polymer structure and a weighted directed message passing neural network (wD-MPNN) architecture that learns specialized representations of molecular ensembles for polymer property prediction. To achieve this, we rely on a parametric description of the underlying distribution of molecules that captures its expectation (i.e., the average graph structure of the repeating unit). We test our model on a new dataset of simulated electronic properties of alternating, random, and block copolymers, and achieve superior performance over graph-based representations that only capture monomeric units as well as robust fingerprint-based ML models. We furthermore evaluate the wD-MPNN on an experimental dataset[50] in which the supervised task involves predicting the possible phases of diblock copolymers. In both tasks, we demonstrate that the explicit inclusion of information about aspects of the molecular ensemble like monomer stoichiometries, chain architectures, and average sizes into the network architecture results in improved predictive performance.

## Methods

In this section, we first review existing cheminformatics approaches used for polymer structure-property regression, which constitute the baseline representations and models we will benchmark our approach against. We then introduce our proposed graph-based representation of polymers, and a graph neural network (GNN) architecture that uses this representation as input. Finally, we describe the new dataset of computed copolymer properties we used to evaluate both traditional and proposed ML approaches.

### *Prior work on polymer representations as model baselines*

Among the representations often used in polymer informatics are molecular fingerprints, which encode the presence or absence of chemical substructures in a binary vector. This representation is directly applicable to a polymer's repeating unit, although it cannot distinguish between isomeric sequences or different monomer stoichiometries. Stoichiometry can be considered by taking the sum of monomer fingerprints, weighted by the respective ratios[29,54]. Alternatively, count fingerprints, which use vectors of integer values and capture the frequency of different chemical patterns, can be applied to oligomeric molecules constructed in a way to reflect the monomers' stoichiometry. By constructing a short polymer chain, the resulting count fingerprints also capture aspects of the polymer's chain architecture. Note this is only partially true when using binary fingerprints. For instance, a random AB copolymer might have specific patterns that identify A-A, A-B, and B-B connections. A block copolymer will have the same patterns, but A-A and B-B will be more frequent than A-B ones. This frequency difference can be captured by count fingerprints, not by binary ones.

A natural representation for small organic molecules, including the repeating units of synthetic polymers, are molecular graphs in which atoms are represented by the graph vertices and bonds by its edges. GNNs[55] take such representation as input to predict molecular properties[56–62], and have been applied to polymer property prediction by considering the structure of individual monomers[27,39]. However, standard GNN architectures cannot handle the inherent stochasticity of polymer structures, as they generally model a specific molecule rather than ensembles. While modeling individual monomeric units may be sufficient for homopolymers, in particular linear ones obtained by chain-growth, predicting properties of copolymers require the ability to distinguish between the constitutional isomers resulting in different chain architectures and sequence ensembles.

Mohapatra et al.[63] have presented a coarse-grained graph representation for macromolecules, which can capture complex macromolecular topologies. Patel et al.[54] have also explored a number of polymer representations, including a similar coarse-grained sequence graph representation. These graph representations can distinguish significantly different macromolecular topologies, but coarse-graining masks information on how monomeric units are bonded to each other. Atomic-level modeling of polymer structure is needed to capture the structure of the connection, which differentiates between structural (e.g., cis versus trans bonds, ortho versus meta substitutions) and sequence (e.g., head-to-tail versus head-to-head or tail-to-tail) isomers. Structural isomers can have vastly different properties. For instance, trans-1,4-polyisoprene (gutta-percha) has a regular structure that allows crystallization and results in a rigid material, while cis-1,4-polyisoprene (natural rubber) is amorphous and elastic. Sequence isomerism can be important instead for polymers synthesized via a step-growth or cationic mechanism, in which the fraction of head-to-tail arrangement can vary based on reactivity and lead to significant differences in polymer properties.

In this work, we adopted both fingerprint- and graph-based representations as baselines approaches. These include Chemprop[64], an established GNN, and random forest (RF) models trained on fingerprint representations. More specifically, Chemprop uses a directed message passing neural network (D-MPNN), a special type of GNN architecture described in more detail later in the Methods. The input for this model was a disconnected graph of the separate monomeric units. The RF models used Extended-Connectivity Fingerprints (ECFP)[65] as input representation. We tested both binary and count fingerprints, constructed from the monomeric units alone, as well as from an ensemble of oligomeric sequences sampled uniformly at random. In the latter case, we sampled up to 32 octameric sequences while satisfying the stoichiometry and chain architecture of the polymer, computed fingerprints for all resulting oligomers, and averaged them to obtain the input representation. We are not aware of prior work using this sequence sampling approach, but we found it to be the most competitive fingerprint-based baseline. More details of the baseline approaches tested are in the SI Extended Methods.

## Graph-based representation of molecular ensembles

Our goal was to expand the architecture of current ML models to capture (i) the recurrent nature of polymers' repeating units, (ii) the different topologies and isomerisms of polymer chains, and (iii) their varying monomer composition and stoichiometry. We thus decided to expand molecular graph representations by incorporating "stochastic" edges to describe the average structure of the repeating unit. In effect, these stochastic edges are bonds weighted by their probability of occurring in the polymer chain (Figure 1).

In our polymer graph representation, each edge is associated with a weight, $w \in (0, 1]$, according to the probability (or frequency) of the bond being present in each repeating unit. By linking separate monomers with edges where $w \leq 1$, we can capture the recurrent nature of polymer chains as well as the ensemble of possible topologies. Figure 1a shows such examples for, e.g., alternating, random, and block copolymers with different sequence isomerisms. For homopolymers and simple alternating copolymers where all edges have a weight of one, this representation naturally reduces to a standard graph representation in which the two ends of the repeating unit have been linked. The periodic representation for crystalline materials proposed by Xie and Grossman[66] is also a special case of the ensemble graph representation proposed here.

Directed edges are necessary to handle a more general set of polymer and oligomer topologies than undirected edges can alone (Figure 1b). Although termini might not exert a strong influence over an overall property of the polymeric material, they provide an apt example to highlight the circumstances that require directed edges. Graph networks learn a hidden representation for each atom in the system based on their neighbors and associated edges. Atoms that connect repeating units mostly have atoms from other repeating units as neighbors, and only infrequently will be connected to the termini. However, atoms that are part of the termini and that connect to the repeating unit always have the repeating unit atoms as neighbors. This asymmetry is needed for a graph network to correctly consider the typical neighborhood of each atom. Some examples of polymer architectures that also require this edge asymmetry are shown in Figure 1b. In graft copolymers, for instance, where the main chain is not fully saturated, the connecting atom on the main chain does not always link to the side-chain polymer, while the connecting atom on the side chain always links to the main chain.

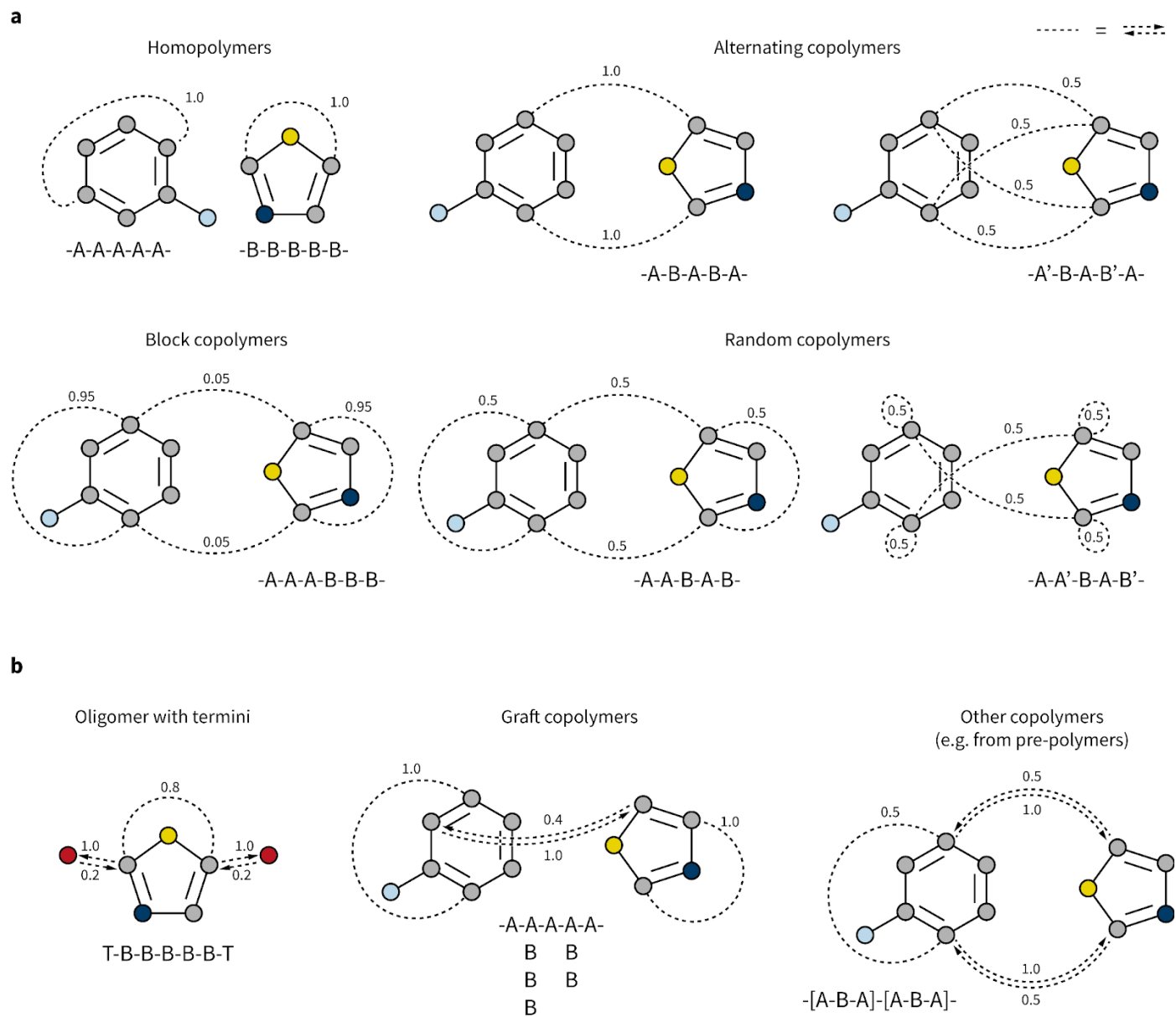

**Figure 1 | Graph representation of selected polymer topologies.** (a) Homopolymers, as well as most alternating, random, and block copolymers can be described as undirected graphs with some stochastic edges, where the probability of the edge reflects the frequency with which the bond is present in the polymer chain. (b) Directed edges enable the representation of a broader range of polymer topologies as graphs, and to capture the effect of termini when needed. Directed edges are necessary when two atoms have different probabilities of being neighbors of each other.

## Graph neural network architecture

The network architecture developed is an extension of the D-MPNN known as Chemprop[64]. MPNNs are a class of GNNs that perform convolutions on graphs while maintaining node-order invariance, and have found broad application for molecular property prediction[58,67–69]. The input of these models is the molecular graph, in which atoms are represented by the graph vertices and bonds by the edges. All nodes and edges are associated with feature vectors that describe the atoms and bonds they represent. These feature vectors are updated iteratively via localized convolutions ("message passing" in the more general framework of MPNNs) that involve neighboring atoms and bonds, and result in learnt embeddings for all nodes and edges. A representation for the whole molecule is then obtained by aggregating (e.g., summing) all atom embeddings. This numerical representation of fixed size is then used by a feed-forward neural network to predict the property of interest, and the whole architecture is trained end to end. Because convolutions and all other operations that manipulate the features of the input graph depend on learnable parameters that are updated by gradient descent, the network is encouraged to learn hidden node, edge, and molecular representations that are highly informative for the predictive task at hand.

In D-MPNNs, the messages used to iteratively update feature vectors are associated with directed edges (bonds), rather than with nodes (atoms) as in regular MPNN architectures[64,70,71]. In addition to having shown state-of-the-art performance on molecular property prediction tasks[64], directed edges are needed for general graph representations of polymers, as discussed above. Here, we propose to weigh directed edges according to their probability of occurring in the polymer chain. As such, we refer to this graph neural network as a weighted D-MPNN (wD-MPNN). The input provided to the wD-MPNN is the graph of the repeating unit of the polymer, in which each node and edge are associated with a set of atom and bond features, $x_v$ and $e_{uv}$, respectively (Figure 2a; details of these features are in the SI Extended Methods).

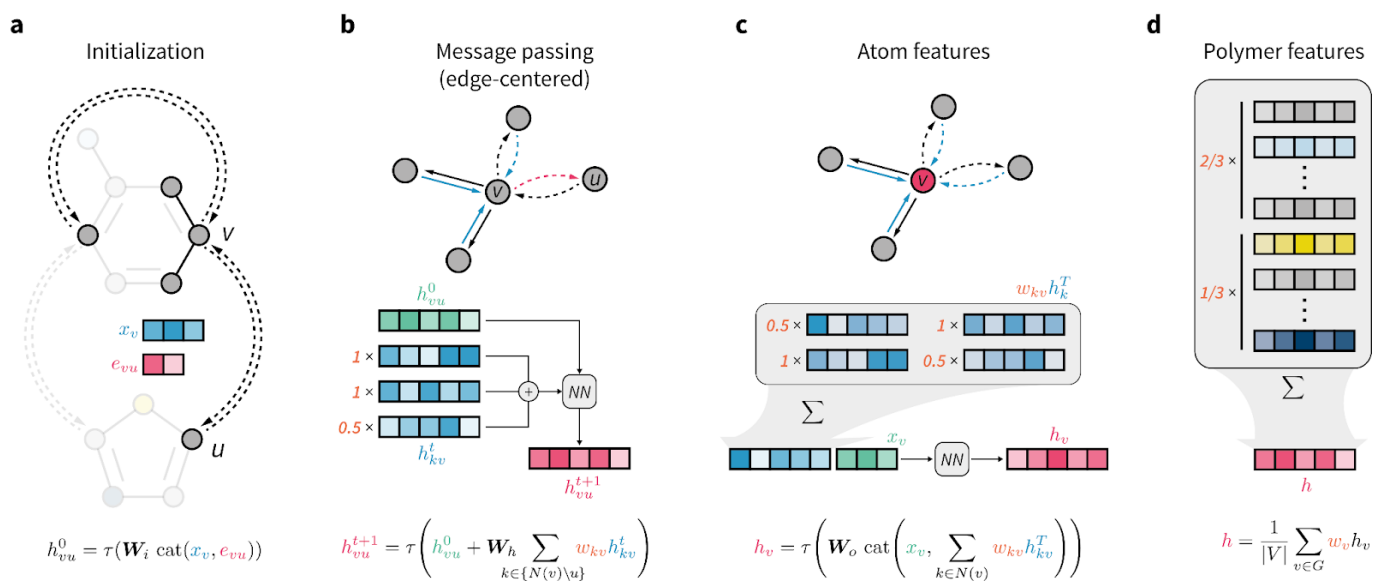

**Figure 2 | Architecture of the directed message passing neural network for polymer property prediction.** (a) Node and edge features are initialized based on corresponding atomic and bond properties, concatenated and passed through a single neural network layer. (b) Message passing is performed for *T* steps, in which edge-centered messages of *v*-outgoing edges are updated based on *v*-incoming edges. Each message is weighted according to user-specified bond probabilities that reflect the topology of the polymer repeating unit. (c) Updated atom features are obtained by a weighted sum over the features of all *v*-incoming edges, followed by concatenation with the initial atom features, and transformation via a single neural network layer. (d) Overall polymer features are obtained by aggregating all final atomic features via a weighted sum or average, where the weights reflect the relative abundance of different substructures (e.g., monomers) in the polymer.

A D-MPNN with messages centered on edges learns a hidden representation $h_{vu}$ for each edge in the graph (Figure 2b). After message passing, a hidden representation for each atom $h_v$ is obtained by considering all of its incoming edges (Figure 2c). In the wD-MPNN, we weigh each edge message according to its probability of being present in the repeating unit, $w_{kv}$, both when updating edge and atom representations (Figure 2b and 2c). Similarly, in existing D-MPNNs, an overall molecular representation $h$ is obtained by averaging or summing over all atom representations $h_v$. In the wD-MPNN, we weigh each $h_v$ according to the relative abundance (i.e., stoichiometry) of the monomer they belong to (Figure 2d) to obtain an overall polymer representation $h$. The aim of incorporating weighted "stochastic" edges and stoichiometry information into the wD-MPNN is to capture a polymer's chain architecture and sequence isomerism by describing its average repeating unit.

The result of the wD-MPNN's processing of the input graph is $h$, a learned numerical representation of the molecular ensemble that defines a polymer and its properties. This is used as the input of a feed-forward neural network to predict the polymer properties of interest, with the whole architecture being trained end-to-end. Additional details of the wD-MPNN architecture are in the SI Extended Methods, and an implementation is available on GitHub (see Data Availability).

*Copolymer dataset*
Given the limited amount of publicly-available polymer data with broad coverage of monomer chemistries, chain architectures, and stoichiometries, we built such a dataset via computation. We considered the chemical space defined by Bai *et al*.[23] comprising conjugated polymers as photocatalysts for hydrogen production (Figure 3a). The full set of possible monomer combinations provides 9 × 862 = 7758 possible co-polymer compositions. In addition to monomer composition, we considered three chain architectures (alternating, random, and block), and three stoichiometric ratios of monomers (1:1, 1:3, 3:1). For random and block copolymers, all ratios were considered, while for perfectly alternating copolymers only the 1:1 stoichiometry was considered. In total, this setup constitutes a space of 42,966 possible copolymers (Figure 3a).

We took electron affinity (EA) and ionization potential (IP) as the properties to be predicted, and generated ground truth labels by computing these properties with density functional tight-binding methods[72]. Specifically, we followed the protocol proposed by Wilbraham and colleagues[40], which involves the computation of EA and IP on oligomers (octamers) with xTB[73], followed by a linear correction based on a calibration against density functional theory (DFT) calculations that used B3LYP density functional[74–77] and the DZP basis-set[78]. For each copolymer, we generated up to 32 sequences and 8 conformers per sequence. In fact, not only random, but also alternating and block copolymers may have multiple possible sequences given the asymmetry of the B monomers, which can result in sequence isomerism. The IP and EA values were Boltzmann averaged across the 8 conformers at 298.15 K, and then averaged across all sequences associated with a specific copolymer (further details in the SI Extended Methods). Ultimately, this process led to a dataset of 42,966 copolymers with different chain architectures and stoichiometric ratios, each labeled with IP and EA values calculated as averages over the ensemble of sequences and conformations.

Both EA and IP were considerably affected by the varying monomer chemistry, chain architecture, and monomer stoichiometry (Figure 2b). Overall, however, monomer chemistry and stoichiometry had a larger impact on EA and IP than chain architecture. Note that an overlapping property distribution, like that shown in Figure 2b for the IP of polymers with different chain architectures (given a specific monomer A and stoichiometry), does not also imply no variation across chain architectures. While the overall distributions overlap, and while IP variation might be smaller than for varying monomer compositions and stoichiometries, the IP is still likely to be different between alternating, random, and block polymer sequences.

In addition to the dataset described above, we created two derivative datasets by artificially inflating the importance of (i) chain architecture and (ii) monomer stoichiometry in determining EA and IP. In the first case, given a specific monomer pair and stoichiometry, the standard deviation of EA and IP values was increased by a factor of 5 while maintaining their original mean values. In the second case, the standard deviation of EA and IP values was increased by a factor of 5 for each specific combination of monomer pairs and chain architecture. These artificial datasets were created to highlight how specific attributes of the wD-MPNN architecture capture property changes directly attributable to different chain architectures and stoichiometries.

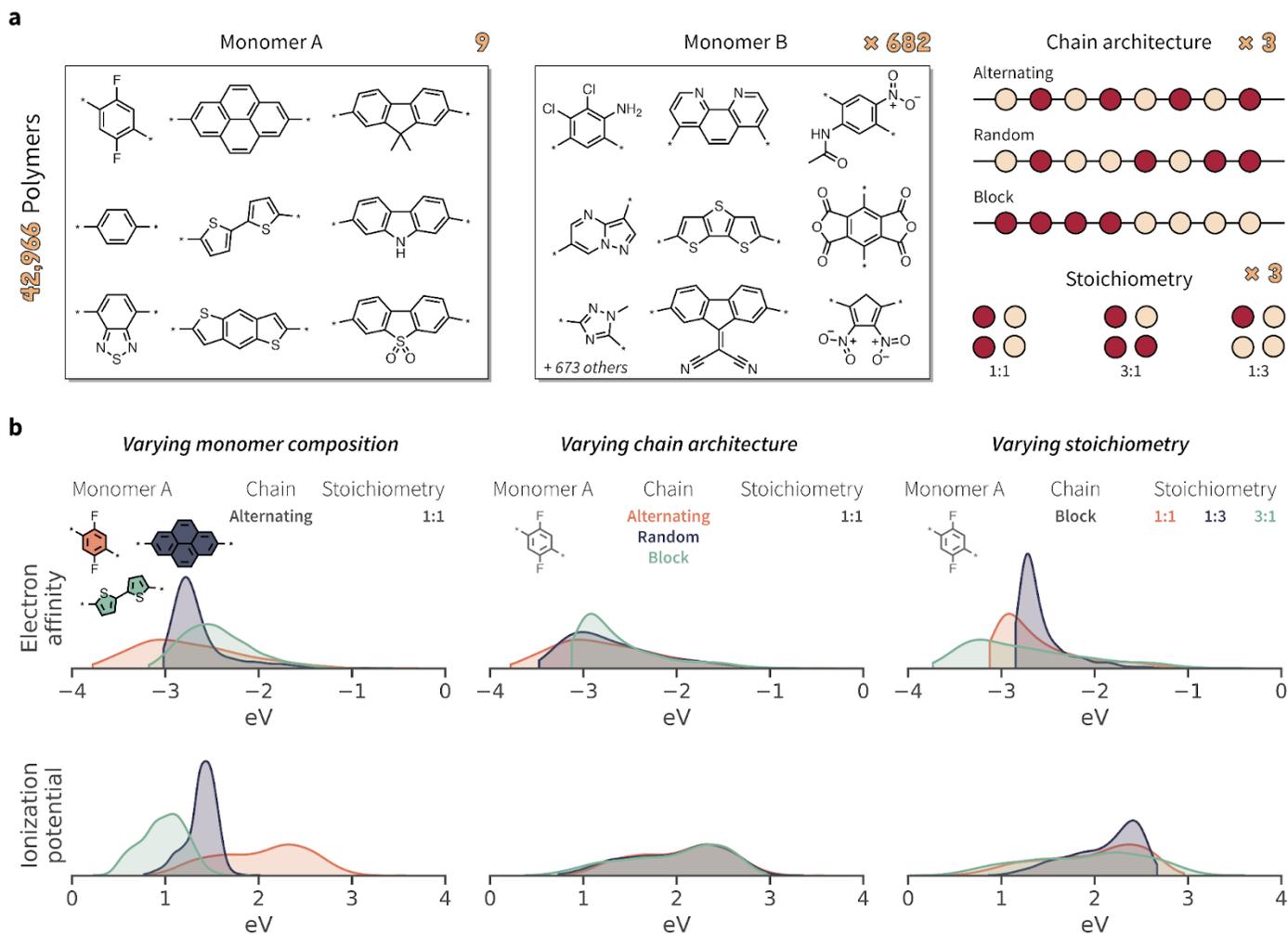

**Figure 3 | Polymers and properties of the dataset.** (a) Building blocks, chain architectures, and stoichiometries present in the dataset of 42,966 copolymers. (b) Probability distributions of electron affinity (EA) and ionization potential (IP) for selected subsets of the copolymer data. The three columns highlight how the property distributions are affected by varying monomer compositions, chain architectures, and stoichiometries. Different monomer chemistry and stoichiometries have the largest impact on EA and IP, while chain architecture had a smaller effect overall. Probability distributions are shown as kernel density estimates, truncated at the respective largest and smallest values.

## Results

In the following sections, we present results in which we tested the ability of our wD-MPNN to predict EA and IP across monomer compositions, stoichiometries, and chain architectures, using our newly-built copolymer dataset. We compared the predictive power of this model against that of baselines models that included a D-MPNN[64] and RF models based on fingerprints, and observed significant improvements in predictive ability over all baselines. We show how, contrary to traditional molecular representations and ML models, the wD-MPNN can discriminate between polymers with the same monomer composition, but different chain architectures and/or monomer stoichiometries. Finally, we demonstrate the use of this model for the prediction of diblock copolymer phases using a recently curated dataset[50].

### Learned polymer representations provide improved predictive performance

The wD-MPNN architecture achieved higher performance than all other models tested on a random 10-fold cross validation split (Figure 4a), saturating the performance measures used with an average coefficient of determination ($R^2$) of 1.00 and a root-mean-square error (RMSE) of 0.03 eV, for both EA and IP predictions. The standard error of the mean was less than 0.005 for both $R^2$ and RMSE. In our discussion of the results, when uncertainty is not provided, we imply it is less than half of the last significant digit.

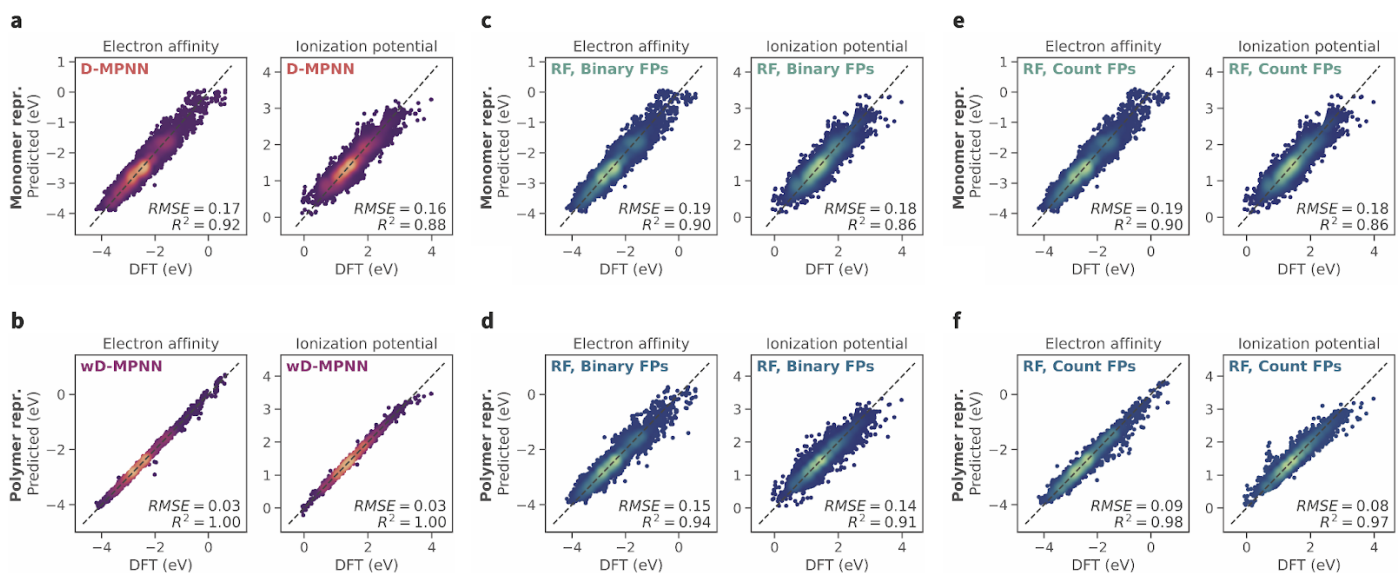

**Figure 4 | Performance of the wD-MPNN and baseline models for the prediction of electron affinity (EA) and ionization potential (IP).** Each parity plot shows the computed DFT values against the ones predicted by the ML models. The parity line is shown as a black dashed line. The scatter/density plots display the predictions of each model for all folds of the 10-fold cross validation. The average coefficient of determination ($R^2$) and root-mean-square error (RMSE, in eV) across all folds are shown; the standard error of the mean is not explicitly shown as it is implied as being less than half of the last significant digit used. (a) Performance of the baseline D-MPNN model, which used a graph representation of the monomeric units as input. (b) Performance of the wD-MPNN model, which is augmented with information about chain architecture and monomer stoichiometry. (c) Performance of a RF model that used a binary fingerprint (FP) representation of the monomeric units as input. (d) Performance of a RF model that used a binary fingerprint representation of the polymer as input, which was obtained as the average fingerprint of a set of oligomeric sequences sampled uniformly at random, while satisfying the correct stoichiometry and chain architecture of the specified polymer. (e) Performance of a RF model that used a count fingerprint representation of the monomeric units as input. (f) Performance of a RF model that used a count fingerprint representation of the polymer as input.

The baseline D-MPNN model achieved RMSEs roughly six times larger (0.17 and 0.16 eV for EA and IP) than those achieved by the wD-MPNN. The RF models that relied on fingerprints representations of the monomeric units returned a performance only marginally inferior to that of the baseline D-MPNN (Figure 4c). RF models with both binary and count fingerprints achieved RMSEs of 0.19 eV and 0.18 eV, for EA and IP, respectively. This performance improved substantially when using averaged fingerprints based on sampled oligomer sequences, which better capture chain architecture and monomer stoichiometry. This was especially true for the RF model using count fingerprints, which achieved RMSEs of 0.09 and 0.08 eV, making it the most competitive baseline approach tested. Despite the excellent performance on this dataset, its RMSE was still three times larger than the one achieved with the wD-MPNN model, and its performance overall qualitatively poorer as visible from the parity plots (Figure 4).

When testing the models on a 9-fold cross-validation where the dataset was split according to the identity of monomer A (Figure 3), performance decreased, as expected (Figure S1). However, the wD-MPNN still achieved RMSEs of 0.10 ± 0.01 and 0.09 ± 0.02 eV, indicating strong generalization performance to new monomer identities. In addition, the performance gap with respect to most other methods increased significantly. The baseline D-MPNN achieved RMSE of 0.20 ± 0.01 and 0.20 ± 0.02 eV. Among the RF models, the highest performance was once again achieved by the representation using averaged count fingerprints across sampled oligomeric sequences, but was considerably worse than that of the D-MPNN models, with RMSEs of 0.25 ± 0.03 and 0.27 ± 0.03 eV, for EA and IP, respectively.

Finally, we tested the data efficiency of the D-MPNN model via multiple, random dataset splits in which we considered training set sizes that included between 43 and 34,373 polymers (i.e., between 0.1% and 80% of the dataset). While the performance of the most competitive RF model (using count fingerprints and sampled polymer chains) was always above that of the baseline D-MPNN, a cross-over point at ~1000 data instances was observed for the wD-MPNN architecture, after which its performance overtook that of RF (Figure S2).

### The wD-MPNN captures how polymer properties depend on chain architecture and monomer stoichiometry

The improved performance of the novel wD-MPNN architecture is a direct result of its ability to discriminate between polymers comprised of the same monomeric units, but differing in their relative abundance (i.e., different stoichiometry) and how they connect to one another to form different sequence ensembles (i.e., different chain architecture). To demonstrate how this information is captured by the additional terms (Methods and Figure 2) provided as inductive biases to the model, we performed ablation studies in which weighted bond information or the terms relative to stoichiometry information were not provided. This resulted in a set of four models: (i) the baseline D-MPNN that is aware of the structure of the separate monomers only, (ii) a D-MPNN that is provided with information of how the monomers may connect to one another to form a chain with specific architecture (alternating, block, or random; information used in the steps shown in Figure 2b and 2c), (iii) a D-MPNN that is provided with information about monomer stoichiometry (information used in the step shown in Figure 2d), and (iv) the full wD-MPNN architecture that is provided with both chain architecture and stoichiometry information.

As discussed above, the baseline D-MPNN model achieved an RMSE of 0.16 eV in the cross-validated prediction of ionization potential (IP). When providing the D-MPNN with information on chain architecture, a small but statistically significant improvement in RMSE was observed, to 0.15 eV. A more substantial improvement was instead observed when the D-MPNN was provided with information on monomer stoichiometries (RMSE = 0.07 eV). This result may have been anticipated given that, overall, monomer stoichiometry was observed to have a larger impact on EA and IP than chain architecture (Figure 3). Yet, both information on stoichiometry and chain architecture was needed by the default D-MPNN to achieve the highest performance (RMSE = 0.03 eV). Equivalent results were obtained also for EA, both when using RMSE and $R^2$ as performance measures, and are reported in Table S1.

While for the specific properties studied here (EA and IP) stoichiometry seemed more important than chain architecture, this is not necessarily the case for other polymer properties. To demonstrate how capturing chain architecture is important in such cases, and to further demonstrate how the wD-MPNN is able to exploit this additional information to achieve superior performance, we created two additional fictitious polymer datasets. These were obtained by artificially inflating the importance of chain architecture and monomer stoichiometry in determining EA and IP (see Methods). While these datasets do not reflect any specific polymeric property, and so we focus on evaluation only in terms of $R^2$, they provide realistic scenarios in which we can control the relative importance of chain architecture and stoichiometry. When chain architecture was made the primary variable determining the IP values, taking this information into account provided the largest performance boost with respect to the baseline model ($R^2$ from 0.65 to 0.86; Table 1). Conversely, when stoichiometry was made artificially even more important, models that did not take it into account could not achieve $R^2$ values above 0.27, while those that did achieved $R^2$ equal or above 0.97. Importantly, in both cases, in which either chain architecture or stoichiometry provided only minimal information, the full wD-MPNN model was able to focus on the most important of the two and always achieve the highest performance of all models tested ($R^2$ of 0.98 and 0.99).

## Predicting diblock copolymer phases from the polymers' chemistry

We further evaluated the wD-MPNN architecture on an experimental dataset that has been recently compiled by Arora *et al*.[50]. This dataset provides the phase behavior of 50 diblock copolymers corresponding to a set of 32 homopolymers and copolymers. It reports the observed copolymer phases (lamellae, hexagonal-packed cylinders, body-centered cubic spheres, acubic gyroid, or disordered) for various relative volume fractions and molar masses, for a total of 4,780 entries. Each entry may be associated with more than one phase, such that the task can be defined as a binary multi-label classification task with five labels, one for each of the phases that can be observed.

The wD-MPNN model was provided with the monomer graphs for both blocks, how these may connect to each other via stochastic edges, and the mole fraction of each block (Figure 2). Here, we also provided the overall copolymer size by scaling the molecular embeddings $h$ by $1 + log(N)$, where $N$ is the degree of polymerization. The scaling factor thus has no effect for chain lengths of one, reducing naturally to the default D-MPNN implementation.

**Table 1 | Effect of capturing chain architecture and monomer stoichiometry information on D-MPNN performance.** The average $R^2$ values obtained from a 10-fold cross validation based on random splits, for the prediction of IP, are shown. Uncertainty is implied as the standard error of the mean was <0.005 in all cases. Under the header "Representation", "monomers" indicates the model was provided with the graph structure of separate monomer units; "chain architecture" indicates the model was provided with information on how the monomer units may connect to one another to form an ensemble of possible sequences, via the definition of edge weights, used as shown in Figure 2b and 2c; "stoichiometry" indicates the model was provided with information on monomer stoichiometry, which was used to weigh learnt node representations as shown in Figure 2d. An extended version of this table, with results obtained also for EA and showing RMSE too as performance measure, is available in Table S1.

| | Representation | | | |
|---|---|---|---|---|
| **Datasets** | Monomers | Monomers + Chain architecture | Monomers + Stoichiometry | Monomers + Chain architecture + Stoichiometry |
| Original dataset | 0.88 | 0.90 | 0.98 | 1.00 |
| Inflated chain architecture importance | 0.65 | 0.86 | 0.71 | 0.98 |
| Inflated stoichiometry importance | 0.26 | 0.27 | 0.97 | 0.99 |

Overall, the wD-MPNN achieved a classification performance, as measured by the area under the precision-recall curve (PRC)[79], of 0.68 ± 0.01 in a 10-fold cross-validation based on stratified splits (Figure 5). Given the imbalance of the five labels, the PRC of a random classifier is expected to be 0.23. When the chain architecture, stoichiometry, and degree of polymerization are not provided to the model, performance drops significantly to a PRC of 0.47 ± 0.01. Considering each of these aspects of the polymer structure improves performance (Figure 5). When information on chain architecture was provided, via weighted edges, the D-MPNN achieved a PRC of 0.49 ± 0.01; when information on polymer size was provided, by scaling molecular embeddings with the degree of polymerization, a PRC of 0.52 ± 0.01 was achieved; and when information on monomer stoichiometry was provided, by scaling atom embeddings with mole fractions, a PRC of 0.67 ± 0.01 was achieved.

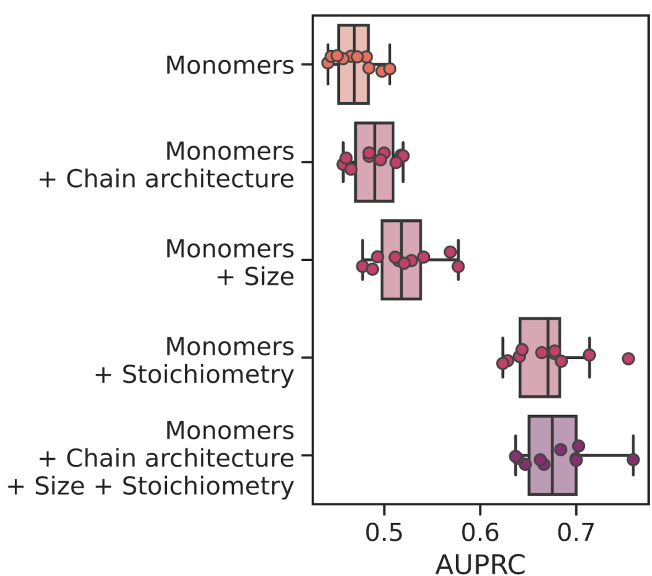

**Figure 5 | Performance of the wD-MPNN and ablated architectures on the classification of diblock copolymer phases.** Performance is measured as the average area under the precision-recall curve (PRC) across the five labels (phases) to be predicted. Each marker reflects the average PRC value, for one fold of a 10-fold cross validation, across the five classification labels. The boxes show the first, second, and third quartiles of the data, and the whiskers extend to up to 1.5 times the interquartile range.

From the results above it emerges how, for this task, the mole fraction of each block is the most informative feature of the polymer. This may be expected given that mole fractions highly correlate with the volume fractions of the two blocks, which is an important factor determining the copolymer phase. In particular, it has been observed that for this dataset very high classification performance can be achieved based *solely* on knowledge of the volume fractions[50]. A RF model trained only on mole fractions achieved a PRC of 0.69 ± 0.01 (Figure S3), and 0.71 ± 0.01 when using solely volume fractions, both of which are better than the wD-MPNN. The highest performance on this task was achieved by a RF model that used count fingerprints with sequence sampling, as well as stoichiometry and size information (PRC of 0.74 ± 0.01; Figure S3). Nevertheless, the relative performance of a structure-based representation in Figure 5 demonstrates the advantages of the wD-MPNN over a monomer-only D-MPNN.

## Discussion

The lack of suitable representations for molecular ensembles is a key obstacle to the development of supervised learning algorithms for polymer informatics. While we have taken a first step toward tailored representations and models for polymer property prediction, these are by no means complete. In particular, the proposed approach only captures the expectation of a molecular ensemble, and not its dispersity[80,81]. The way in which the chain architecture of polymers is described via weighted edges is representative of an average repeating unit. As such, this representation would not distinguish between gradient[82,83] and block copolymers with the same average block length. Similarly, the use of the average chain length alone as a scaling factor for the molecular embeddings results in a model that cannot distinguish between polymers with equal average size but different polydispersity. Graph network architectures that better capture the heterogeneity of molecular ensembles could be explored in the future by, for instance, expanding the parametric approach used in this work to higher-order moments beyond the mean of the distribution of connections. Despite these limitations, the approach developed represents a first step toward better ML representations of materials that are composed of large molecular ensembles.

In the copolymer phase prediction task, we incorporated information on polymer size in the wD-MPNN architecture by scaling the learnt polymer embeddings. However, there are alternative approaches that could be explored with suitable datasets. Another way to incorporate size information explicitly into the model would be to append the degree of polymerization, or the molar mass, to the embedding vector $h$ after message passing. This also provides a general means to have the wD-MPNN consider information about process parameters. However, when information about the termini is available, the weights $w_{kv}$ associated with the stochastic edges of the termini, together with the weights $w_v$ reflecting the stoichiometric ratio between different building blocks, in principle would already capture average chain length implicitly. As more copolymer datasets become available, one could explore multiple ways to integrate size information into the wD-MPNN architecture and study the performance and generality of different approaches.

To further advance these ML models and the field of polymer informatics, data availability is fundamental. Here, we have built and provided a computational dataset of over 40,000 polymers that may be used to further develop tailored ML models for polymer property prediction. Yet, more such datasets are needed to increase the diversity of polymer prediction tasks, each of which might uniquely be affected by different aspects of the ensemble of molecules defining the material. While expensive, properties computed via electronic structure[39,40] or molecular dynamics calculations[84] provide a means to obtain comprehensive and relatively noise-free datasets to establish the first generation of ML models designed specifically for polymers. Despite not yet being readily available, properties obtained via atomistic molecular dynamics simulations may be especially complementary to the dataset provided here, as they may more strongly depend on intramolecular interactions, conformational ensembles, and chain length. In the meantime, it will be important to create open-access databases of experimentally-measured properties available to the community, and in machine readable format, reflecting similar and established initiatives in other chemistry fields[85–93]. Efforts like PolyInfo[94] and Polymer Genome[25] attempt to tackle this challenge, but the data in these databases is not truly open access. Open initiatives that aim at building findable as well as accessible databases, like the Community Resource for Innovation in Polymer Technology (CRIPT)[95], will likely play an increasingly important role in enabling tailored ML models for polymer informatics.

The wD-MPNN model described in this work is particularly useful in polymer design campaigns in which exploring a broad range of monomer chemistries and compositions, chain architectures, and polymer sizes is of interest. When this is not the case, however, and one would like to focus on a small set of monomers and a well-defined chain architecture (e.g., only alternating copolymers, or even sequence-defined polymers), the use of such a model is not necessarily advantageous with respect to more traditional ML models. Indeed, if a ML algorithm is not required to distinguish between polymers with, e.g, different chain architectures, average sizes, or monomer stoichiometries, then the structure of the monomers alone or the use of hand-crafted descriptors will be sufficient. Furthermore, the availability of highly informative descriptors or proxy observables may obviate the need for a deep learning model, as we noticed for the task of predicting the phases of diblock copolymers. Finally, the model choice might also be forced by data availability. As discussed in the Results section, for the task of predicting EA and IP we found that with fewer than ~1000 data instances the wD-MPNN did not provide an advantage over a RF model. Only when >1000 examples were provided for training did the wD-MPNN overtake the performance seen for RF (Figure S2).

## Conclusion

In this work we have developed a graph representation of molecular ensembles and an associated wD-MPNN architecture with immediate relevance to polymer property prediction. We have shown how this approach captures critical features of polymeric materials, like chain architecture, monomer stoichiometry, and expected size, to achieve superior accuracy with respect to baseline approaches that disregard this information. We have furthermore developed competitive baseline models based on random forest, count fingerprints, and sequence sampling. To evaluate the performance of the different ML models, we generated a dataset with electron affinity and ionization potential values for over 40k polymers with varying monomer composition, stoichiometry, and chain architecture via ~15 million single point energy calculations. Both this dataset and the ML models developed constitute a positive step toward next-generation algorithms for polymer informatics and provide an avenue for modeling the properties of molecular ensembles.

## Supporting Information

The Supporting Information is available free of charge.
- Extended Methods, supporting figures and tables (PDF)
- Dataset of computed electron affinity and ionization potential values for 42,966 copolymers (CSV)

## Acknowledgments

M.A. and C.W.C. thank Guangqi Wu, Bradley D. Olsen, Hursh V. Sureka, and Katharina Fransen for useful discussions on polymer chemistry. M.A. thanks Thijs Stuyver for help and discussions related to the electronic structure calculations, Rebecca Neeser for suggestions on data visualization, and John Bradshaw and Samuel Goldman for discussions on the results of this work. Research reported in this publication was supported by NIGMS of the National Institutes of Health under award number R21GM141616. The content is solely the responsibility of the authors and does not necessarily represent the official views of the National Institutes of Health.

## Data availability

The wD-MPNN model developed is available on GitHub at https://github.com/coleygroup/polymer-chemprop (v1.4.0-polymer). The dataset with electron affinity and ionization potential values for 42,966 copolymers is provided as part of the Supporting Information as a CSV file, and on GitHub at https://github.com/coleygroup/polymer-chemprop-data. Jupyter notebooks and Python scripts needed to create the dataset and reproduce the results of this manuscript are also available in the same GitHub repository.

## References

(1) Coleman, M. M.; Painter, P. C. *Fundamentals of Polymer Science: An Introductory Text, Second Edition*; Taylor & Francis, 1998.
(2) McCrum, N. G.; Buckley, C. P.; . Bucknall, C. B.; Bucknall, C. B. *Principles of Polymer Engineering*; Oxford University Press, 1997.
(3) Ruzette, A.-V.; Leibler, L. Block Copolymers in Tomorrow's Plastics. *Nat. Mater.* **2005**, *4* (1), 19–31.
(4) Kataoka, K.; Harada, A.; Nagasaki, Y. Block Copolymer Micelles for Drug Delivery: Design, Characterization and Biological Significance. *Adv. Drug Deliv. Rev.* **2001**, *47* (1), 113–131.
(5) Islam, M. T.; Rahman, M. M.; Mazumder, N.-U.-S. Polymers for Textile Production. *Frontiers of Textile Materials*. Wiley March 4, 2020, pp 13–59. https://doi.org/10.1002/9781119620396.ch2.
(6) Lopez, J.; Mackanic, D. G.; Cui, Y.; Bao, Z. Designing Polymers for Advanced Battery Chemistries. *Nature Reviews Materials* **2019**, *4* (5), 312–330.
(7) Patil, A.; Patel, A.; Purohit, R. An Overview of Polymeric Materials for Automotive Applications. *Materials Today: Proceedings* **2017**, *4* (2, Part A), 3807–3815.
(8) Sung, Y. K.; Kim, S. W. Recent Advances in Polymeric Drug Delivery Systems. *Biomater Res* **2020**, *24*, 12.
(9) Huang, H.-J.; Tsai, Y.-L.; Lin, S.-H.; Hsu, S.-H. Smart Polymers for Cell Therapy and Precision Medicine. *J. Biomed. Sci.* **2019**, *26* (1), 73.
(10) Liechty, W. B.; Kryscio, D. R.; Slaughter, B. V.; Peppas, N. A. Polymers for Drug Delivery Systems. *Annu. Rev. Chem. Biomol. Eng.* **2010**, *1*, 149–173.
(11) Kumar, R.; Le, N.; Tan, Z.; Brown, M. E.; Jiang, S.; Reineke, T. M. Efficient Polymer-Mediated Delivery of Gene-Editing Ribonucleoprotein Payloads through Combinatorial Design, Parallelized Experimentation, and Machine Learning. *ACS Nano* **2020**. https://doi.org/10.1021/acsnano.0c08549.
(12) Kumar, R.; Le, N.; Oviedo, F.; Brown, M. E.; Reineke, T. M. Combinatorial Polycation Synthesis and Causal Machine Learning Reveal Divergent Polymer Design Rules for Effective pDNA and Ribonucleoprotein Delivery. *JACS Au* **2022**, *2* (2), 428–442.
(13) Kumar, R.; Santa Chalarca, C. F.; Bockman, M. R.; Van Bruggen, C.; Grimme, C. J.; Dalal, R. J.; Hanson, M. G.; Hexum, J. K.; Reineke, T. M. Polymeric Delivery of Therapeutic Nucleic Acids. *Chem. Rev.* **2021**, *121* (18), 11527–11652.
(14) Bannigan, P.; Aldeghi, M.; Bao, Z.; Häse, F.; Aspuru-Guzik, A.; Allen, C. Machine Learning Directed Drug Formulation Development. *Adv. Drug Deliv. Rev.* **2021**, *175*, 113806.
(15) Tan, Z.; Jiang, Y.; Ganewatta, M. S.; Kumar, R.; Keith, A. Block Polymer Micelles Enable CRISPR/Cas9 Ribonucleoprotein Delivery: Physicochemical Properties Affect Packaging Mechanisms and Gene Editing Efficiency. **2019**.

(16)     Mitchell, M. J.; Billingsley, M. M.; Haley, R. M.; Wechsler, M. E.; Peppas, N. A.; Langer, R. Engineering Precision Nanoparticles for Drug Delivery. *Nat. Rev. Drug Discov.* **2021**, *20* (2), 101–124.
(17)     Obermeyer, A. C.; Mills, C. E.; Dong, X.-H.; Flores, R. J.; Olsen, B. D. Complex Coacervation of Supercharged Proteins with Polyelectrolytes. *Soft Matter* **2016**, *12* (15), 3570–3581.
(18)     Ulbrich, K.; Holá, K.; Šubr, V.; Bakandritsos, A.; Tuček, J.; Zbořil, R. Targeted Drug Delivery with Polymers and Magnetic Nanoparticles: Covalent and Noncovalent Approaches, Release Control, and Clinical Studies. *Chem. Rev.* **2016**, *116* (9), 5338–5431.
(19)     Bannigan, P.; Bao, Z.; Hickman, R.; Aldeghi, M.; Häse, F.; Aspuru-Guzik, A.; Allen, C. Machine Learning Models to Accelerate the Design of Polymeric Long-Acting Injectables. *ChemRxiv* **2022**. https://doi.org/10.26434/chemrxiv-2021-mxrxw-v2.
(20)     Mayer, A. C.; Scully, S. R.; Hardin, B. E.; Rowell, M. W.; McGehee, M. D. Polymer-Based Solar Cells. *Mater. Today* **2007**, *10* (11), 28–33.
(21)     Lee, C.; Lee, S.; Kim, G.-U.; Lee, W.; Kim, B. J. Recent Advances, Design Guidelines, and Prospects of All-Polymer Solar Cells. *Chem. Rev.* **2019**, *119* (13), 8028–8086.
(22)     Muench, S.; Wild, A.; Friebe, C.; Häupler, B.; Janoschka, T.; Schubert, U. S. Polymer-Based Organic Batteries. *Chem. Rev.* **2016**, *116* (16), 9438–9484.
(23)     Bai, Y.; Wilbraham, L.; Slater, B. J.; Zwijnenburg, M. A.; Sprick, R. S.; Cooper, A. I. Accelerated Discovery of Organic Polymer Photocatalysts for Hydrogen Evolution from Water through the Integration of Experiment and Theory. *J. Am. Chem. Soc.* **2019**, *141* (22), 9063–9071.
(24)     Wang, Y.; Vogel, A.; Sachs, M.; Sprick, R. S.; Wilbraham, L. Current Understanding and Challenges of Solar-Driven Hydrogen Generation Using Polymeric Photocatalysts. *Nature Energy* **2019**.
(25)     Kim, C.; Chandrasekaran, A.; Huan, T. D.; Das, D.; Ramprasad, R. Polymer Genome: A Data-Powered Polymer Informatics Platform for Property Predictions. *J. Phys. Chem. C* **2018**, *122* (31), 17575–17585.
(26)     Jha, A.; Chandrasekaran, A.; Kim, C.; Ramprasad, R. Impact of Dataset Uncertainties on Machine Learning Model Predictions: The Example of Polymer Glass Transition Temperatures. *Modell. Simul. Mater. Sci. Eng.* **2019**, *27* (2), 024002.
(27)     Tao, L.; Varshney, V.; Li, Y. Benchmarking Machine Learning Models for Polymer Informatics: An Example of Glass Transition Temperature. *J. Chem. Inf. Model.* **2021**, *61* (11), 5395–5413.
(28)     Park, J.; Shim, Y.; Lee, F.; Rammohan, A.; Goyal, S.; Shim, M.; Jeong, C.; Kim, D. S. Prediction and Interpretation of Polymer Properties Using the Graph Convolutional Network. *ACS Polym. Au* **2022**. https://doi.org/10.1021/acspolymersau.1c00050.
(29)     Kuenneth, C.; Schertzer, W.; Ramprasad, R. Copolymer Informatics with Multitask Deep Neural Networks. *Macromolecules* **2021**, *54* (13), 5957–5961.
(30)     Tao, L.; Chen, G.; Li, Y. Machine Learning Discovery of High-Temperature Polymers. *Patterns* **2021**, *2* (4), 100225.
(31)     Wu, S.; Kondo, Y.; Kakimoto, M.-A.; Yang, B.; Yamada, H.; Kuwajima, I.; Lambard, G.; Hongo, K.; Xu, Y.; Shiomi, J.; Schick, C.; Morikawa, J.; Yoshida, R. Machine-Learning-Assisted Discovery of Polymers with High Thermal Conductivity Using a Molecular Design Algorithm. *npj Computational Materials* **2019**, *5* (1), 1–11.
(32)     Yamada, H.; Liu, C.; Wu, S.; Koyama, Y.; Ju, S.; Shiomi, J.; Morikawa, J.; Yoshida, R. Predicting Materials Properties with Little Data Using Shotgun Transfer Learning. *ACS Cent. Sci.* **2019**, *5* (10), 1717–1730.
(33)     Venkatram, S.; Kim, C.; Chandrasekaran, A.; Ramprasad, R. Critical Assessment of the Hildebrand and Hansen Solubility Parameters for Polymers. *J. Chem. Inf. Model.* **2019**, *59* (10), 4188–4194.
(34)     Wang, M.; Xu, Q.; Tang, H.; Jiang, J. Machine Learning-Enabled Prediction and High-Throughput Screening of Polymer Membranes for Pervaporation Separation. *ACS Appl. Mater. Interfaces* **2022**, *14* (6), 8427–8436.
(35)     Barnett, J. W.; Bilchak, C. R.; Wang, Y.; Benicewicz, B. C.; Murdock, L. A.; Bereau, T.; Kumar, S. K. Designing Exceptional Gas-Separation Polymer Membranes Using Machine Learning. *Science Advances* **2020**, *6* (20), eaaz4301.
(36)     Wilbraham, L.; Sprick, R. S.; Jelfs, K. E.; Zwijnenburg, M. A. Mapping Binary Copolymer Property Space with Neural Networks. *Chem. Sci.* **2019**, *10* (19), 4973–4984.
(37)     Patra, A.; Batra, R.; Chandrasekaran, A.; Kim, C.; Huan, T. D.; Ramprasad, R. A Multi-Fidelity Information-Fusion Approach to Machine Learn and Predict Polymer Bandgap. *Comput. Mater. Sci.* **2020**, *172*, 109286.
(38)     Doan Tran, H.; Kim, C.; Chen, L.; Chandrasekaran, A.; Batra, R.; Venkatram, S.; Kamal, D.; Lightstone, J. P.; Gurnani, R.; Shetty, P.; Ramprasad, M.; Laws, J.; Shelton, M.; Ramprasad, R. Machine-Learning Predictions of Polymer Properties with Polymer Genome. *J. Appl. Phys.* **2020**, *128* (17), 171104.
(39)     St. John, P. C.; Phillips, C.; Kemper, T. W.; Wilson, A. N.; Guan, Y.; Crowley, M. F.; Nimlos, M. R.; Larsen, R. E. Message-Passing Neural Networks for High-Throughput Polymer Screening. *J. Chem. Phys.* **2019**, *150* (23), 234111.

(40)     Wilbraham, L.; Berardo, E.; Turcani, L.; Jelfs, K. E.; Zwijnenburg, M. A. High-Throughput Screening Approach for the Optoelectronic Properties of Conjugated Polymers. *J. Chem. Inf. Model.* **2018**, *58* (12), 2450–2459.
(41)     Mannodi-Kanakkithodi, A.; Pilania, G.; Huan, T. D.; Lookman, T.; Ramprasad, R. Machine Learning Strategy for Accelerated Design of Polymer Dielectrics. *Sci. Rep.* **2016**, *6* (1), 1–10.
(42)     Chen, L.; Kim, C.; Batra, R.; Lightstone, J. P.; Wu, C.; Li, Z.; Deshmukh, A. A.; Wang, Y.; Tran, H. D.; Vashishta, P.; Sotzing, G. A.; Cao, Y.; Ramprasad, R. Frequency-Dependent Dielectric Constant Prediction of Polymers Using Machine Learning. *npj Computational Materials* **2020**, *6* (1), 1–9.
(43)     Reis, M.; Gusev, F.; Taylor, N. G.; Chung, S. H.; Verber, M. D.; Lee, Y. Z.; Isayev, O.; Leibfarth, F. A. Machine-Learning-Guided Discovery of 19F MRI Agents Enabled by Automated Copolymer Synthesis. *J. Am. Chem. Soc.* **2021**, *143* (42), 17677–17689.
(44)     Kuenneth, C.; Rajan, A. C.; Tran, H.; Chen, L.; Kim, C.; Ramprasad, R. Polymer Informatics with Multi-Task Learning. *Patterns Prejudice* **2021**, *2* (4), 100238.
(45)     Chen, L.; Pilania, G.; Batra, R.; Huan, T. D.; Kim, C.; Kuenneth, C.; Ramprasad, R. Polymer Informatics: Current Status and Critical next Steps. *Mater. Sci. Eng. R Rep.* **2021**, *144*, 100595.
(46)     Wu, S.; Yamada, H.; Hayashi, Y.; Zamengo, M.; Yoshida, R. Potentials and Challenges of Polymer Informatics: Exploiting Machine Learning for Polymer Design. **2020**.
(47)     Audus, D. J.; de Pablo, J. J. Polymer Informatics: Opportunities and Challenges. *ACS Macro Lett.* **2017**, *6* (10), 1078–1082.
(48)     Peerless, J. S.; Milliken, N. J. B.; Oweida, T. J.; Manning, M. D.; Yingling, Y. G. Soft Matter Informatics: Current Progress and Challenges. *Adv. Theory Simul.* **2019**, *2* (1), 1800129.
(49)     Patra, T. K. Data-Driven Methods for Accelerating Polymer Design. *ACS Polym. Au* **2022**, *2* (1), 8–26.
(50)     Arora, A.; Lin, T.-S.; Rebello, N. J.; Av-Ron, S. H. M.; Mochigase, H.; Olsen, B. D. Random Forest Predictor for Diblock Copolymer Phase Behavior. *ACS Macro Lett.* **2021**, *10* (11), 1339–1345.
(51)     Lin, T.-S.; Coley, C. W.; Mochigase, H.; Beech, H. K.; Wang, W.; Wang, Z.; Woods, E.; Craig, S. L.; Johnson, J. A.; Kalow, J. A.; Jensen, K. F.; Olsen, B. D. BigSMILES: A Structurally-Based Line Notation for Describing Macromolecules. *ACS Cent Sci* **2019**, *5* (9), 1523–1531.
(52)     Chithrananda, S.; Grand, G.; Ramsundar, B. ChemBERTa: Large-Scale Self-Supervised Pretraining for Molecular Property Prediction. *Machine Learning for Molecules Workshop at NeurIPS*. 2020.
(53)     Fabian, B.; Edlich, T.; Gaspar, H.; Segler, M. H. S.; Meyers, J.; Fiscato, M.; Ahmed, M. Molecular Representation Learning with Language Models and Domain-Relevant Auxiliary Tasks. *Machine Learning for Molecules Workshop at NeurIPS*. 2020.
(54)     Patel, R. A.; Borca, C. H.; Webb, M. A. Featurization Strategies for Polymer Sequence or Composition Design by Machine Learning. *Mol. Syst. Des. Eng.* **2022**, -.
(55)     Zhou, J.; Cui, G.; Hu, S.; Zhang, Z.; Yang, C.; Liu, Z.; Wang, L.; Li, C.; Sun, M. Graph Neural Networks: A Review of Methods and Applications. *AI Open* **2020**, *1*, 57–81.
(56)     Duvenaud, D. K.; Maclaurin, D.; Iparraguirre, J.; Bombarell, R.; Hirzel, T.; Aspuru-Guzik, A.; Adams, R. P. Convolutional Networks on Graphs for Learning Molecular Fingerprints. In *Advances in Neural Information Processing Systems*; Cortes, C., Lawrence, N., Lee, D., Sugiyama, M., Garnett, R., Eds.; Curran Associates, Inc., 2015; Vol. 28.
(57)     Kearnes, S.; McCloskey, K.; Berndl, M.; Pande, V.; Riley, P. Molecular Graph Convolutions: Moving beyond Fingerprints. *J. Comput. Aided Mol. Des.* **2016**, *30* (8), 595–608.
(58)     Stokes, J. M.; Yang, K.; Swanson, K.; Jin, W.; Cubillos-Ruiz, A.; Donghia, N. M.; MacNair, C. R.; French, S.; Carfrae, L. A.; Bloom-Ackermann, Z.; Tran, V. M.; Chiappino-Pepe, A.; Badran, A. H.; Andrews, I. W.; Chory, E. J.; Church, G. M.; Brown, E. D.; Jaakkola, T. S.; Barzilay, R.; Collins, J. J. A Deep Learning Approach to Antibiotic Discovery. *Cell* **2020**, *181* (2), 475–483.
(59)     Sánchez-Lengeling, B.; Wei, J. N.; Lee, B. K.; Gerkin, R. C.; Aspuru-Guzik, A.; Wiltschko, A. B. Machine Learning for Scent: Learning Generalizable Perceptual Representations of Small Molecules. *ArXiv* **2019**.
(60)     Tsubaki, M.; Tomii, K.; Sese, J. Compound–protein Interaction Prediction with End-to-End Learning of Neural Networks for Graphs and Sequences. *Bioinformatics* **2018**, *35* (2), 309–318.
(61)     Jiang, D.; Wu, Z.; Hsieh, C.-Y.; Chen, G.; Liao, B.; Wang, Z.; Shen, C.; Cao, D.; Wu, J.; Hou, T. Could Graph Neural Networks Learn Better Molecular Representation for Drug Discovery? A Comparison Study of Descriptor-Based and Graph-Based Models. *J. Cheminform.* **2021**, *13* (1), 12.
(62)     McCloskey, K.; Sigel, E. A.; Kearnes, S.; Xue, L.; Tian, X.; Moccia, D.; Gikunju, D.; Bazzaz, S.; Chan, B.; Clark, M. A.; Cuozzo, J. W.; Guié, M.-A.; Guilinger, J. P.; Huguet, C.; Hupp, C. D.; Keefe, A. D.; Mulhern, C. J.; Zhang, Y.; Riley, P. Machine Learning on DNA-Encoded Libraries: A New Paradigm for Hit Finding. *J. Med. Chem.* **2020**, *63* (16), 8857–8866.
(63)     Mohapatra, S.; An, J.; Gómez-Bombarelli, R. Chemistry-Informed Macromolecule Graph Representation for Similarity Computation, Unsupervised and Supervised Learning. *Mach. Learn.: Sci. Technol.* **2022**, *3* (1), 015028.

(64) Yang, K.; Swanson, K.; Jin, W.; Coley, C.; Eiden, P.; Gao, H.; Guzman-Perez, A.; Hopper, T.; Kelley, B.; Mathea, M.; Palmer, A.; Settels, V.; Jaakkola, T.; Jensen, K.; Barzilay, R. Analyzing Learned Molecular Representations for Property Prediction. *J. Chem. Inf. Model.* **2019**, *59* (8), 3370–3388.

(65) Rogers, D.; Hahn, M. Extended-Connectivity Fingerprints. *J. Chem. Inf. Model.* **2010**, *50* (5), 742–754.

(66) Xie, T.; Grossman, J. C. Crystal Graph Convolutional Neural Networks for an Accurate and Interpretable Prediction of Material Properties. *Phys. Rev. Lett.* **2018**, *120* (14), 145301.

(67) Gilmer, J.; Schoenholz, S. S.; Riley, P. F.; Vinyals, O.; Dahl, G. E. Neural Message Passing for Quantum Chemistry. In *Proceedings of the 34th International Conference on Machine Learning - Volume 70*; ICML'17; JMLR.org, 2017; pp 1263–1272.

(68) Heid, E.; Green, W. H. Machine Learning of Reaction Properties via Learned Representations of the Condensed Graph of Reaction. *J. Chem. Inf. Model.* **2021**. https://doi.org/10.1021/acs.jcim.1c00975.

(69) Flam-Shepherd, D.; Wu, T. C.; Friederich, P.; Aspuru-Guzik, A. Neural Message Passing on High Order Paths. *Mach. Learn.: Sci. Technol.* **2021**, *2* (4), 045009.

(70) Dai, H.; Dai, B.; Song, L. Discriminative Embeddings of Latent Variable Models for Structured Data. In *Proceedings of the 33rd International Conference on International Conference on Machine Learning - Volume 48*; ICML'16; JMLR.org, 2016; pp 2702–2711.

(71) Gasteiger, J.; Groß, J.; Günnemann, S. Directional Message Passing for Molecular Graphs. In *International Conference on Learning Representations*; 2020.

(72) Bannwarth, C.; Caldeweyher, E.; Ehlert, S.; Hansen, A.; Pracht, P.; Seibert, J.; Spicher, S.; Grimme, S. Extended Tight–binding Quantum Chemistry Methods. *Wiley Interdiscip. Rev. Comput. Mol. Sci.* **2021**, *11* (2). https://doi.org/10.1002/wcms.1493.

(73) Grimme, S.; Bannwarth, C.; Shushkov, P. A Robust and Accurate Tight-Binding Quantum Chemical Method for Structures, Vibrational Frequencies, and Noncovalent Interactions of Large Molecular Systems Parametrized for All Spd-Block Elements (Z = 1–86). *J. Chem. Theory Comput.* **2017**, *13* (5), 1989–2009.

(74) Vosko, S. H.; Wilk, L.; Nusair, M. Accurate Spin-Dependent Electron Liquid Correlation Energies for Local Spin Density Calculations: A Critical Analysis. *Can. J. Phys.* **1980**, *59*, 1200.

(75) Becke, A. D. Density–functional Thermochemistry. III. The Role of Exact Exchange. *J. Chem. Phys.* **1993**, *98* (7), 5648–5652.

(76) Lee, C.; Yang, W.; Parr, R. G. Development of the Colle-Salvetti Correlation-Energy Formula into a Functional of the Electron Density. *Phys. Rev. B Condens. Matter* **1988**, *37* (2), 785–789.

(77) Stephens, P. J.; Devlin, F. J.; Chabalowski, C. F.; Frisch, M. J. Ab Initio Calculation of Vibrational Absorption and Circular Dichroism Spectra Using Density Functional Force Fields. *J. Phys. Chem.* **1994**, *98* (45), 11623–11627.

(78) Schäfer, A.; Horn, H.; Ahlrichs, R. Fully Optimized Contracted Gaussian Basis Sets for Atoms Li to Kr. *J. Chem. Phys.* **1992**, *97* (4), 2571–2577.

(79) Saito, T.; Rehmsmeier, M. The Precision-Recall Plot Is More Informative than the ROC Plot When Evaluating Binary Classifiers on Imbalanced Datasets. *PLoS One* **2015**, *10* (3), e0118432.

(80) Lynd, N. A.; Hillmyer, M. A. Influence of Polydispersity on the Self-Assembly of Diblock Copolymers. *Macromolecules* **2005**, *38* (21), 8803–8810.

(81) Gentekos, D. T.; Dupuis, L. N.; Fors, B. P. Beyond Dispersity: Deterministic Control of Polymer Molecular Weight Distribution. *J. Am. Chem. Soc.* **2016**, *138* (6), 1848–1851.

(82) Grune, E.; Appold, M.; Müller, A. H. E.; Gallei, M.; Frey, H. Anionic Copolymerization Enables the Scalable Synthesis of Alternating (AB)n Multiblock Copolymers with High Molecular Weight in n/2 Steps. *ACS Macro Lett.* **2018**, *7* (7), 807–810.

(83) Alam, M. M.; Jack, K. S.; Hill, D. J. T.; Whittaker, A. K.; Peng, H. Gradient Copolymers – Preparation, Properties and Practice. *Eur. Polym. J.* **2019**, *116*, 394–414.

(84) Webb, M. A.; Jackson, N. E.; Gil, P. S.; de Pablo, J. J. Targeted Sequence Design within the Coarse-Grained Polymer Genome. *Sci Adv* **2020**, *6* (43). https://doi.org/10.1126/sciadv.abc6216.

(85) Kearnes, S. M.; Maser, M. R.; Wleklinski, M.; Kast, A.; Doyle, A. G.; Dreher, S. D.; Hawkins, J. M.; Jensen, K. F.; Coley, C. W. The Open Reaction Database. *J. Am. Chem. Soc.* **2021**, *143* (45), 18820–18826.

(86) Liu, T.; Lin, Y.; Wen, X.; Jorissen, R. N.; Gilson, M. K. BindingDB: A Web-Accessible Database of Experimentally Determined Protein-Ligand Binding Affinities. *Nucleic Acids Res.* **2007**, *35* (Database issue), D198–D201.

(87) Gilson, M. K.; Liu, T.; Baitaluk, M.; Nicola, G.; Hwang, L.; Chong, J. BindingDB in 2015: A Public Database for Medicinal Chemistry, Computational Chemistry and Systems Pharmacology. *Nucleic Acids Res.* **2015**, *44* (D1), D1045–D1053.

(88) Berman, H. M.; Westbrook, J.; Feng, Z.; Gilliland, G.; Bhat, T. N.; Weissig, H.; Shindyalov, I. N.; Bourne, P. E. The Protein Data Bank. *Nucleic Acids Res.* **2000**, *28* (1), 235–242.

(89) Zardecki, C.; Dutta, S.; Goodsell, D. S.; Voigt, M.; Burley, S. K. RCSB Protein Data Bank: A Resource for Chemical, Biochemical, and Structural Explorations of Large and Small Biomolecules. *J. Chem. Educ.* **2016**, *93* (3), 569–575.
(90) Berman, H. M.; Gierasch, L. M. How the Protein Data Bank Changed Biology: An Introduction to the JBC Reviews Thematic Series, Part 1. *J. Biol. Chem.* **2021**, *296*. https://doi.org/10.1016/j.jbc.2021.100608.
(91) Gaulton, A.; Bellis, L. J.; Bento, A. P.; Chambers, J.; Davies, M.; Hersey, A.; Light, Y.; McGlinchey, S.; Michalovich, D.; Al-Lazikani, B.; Overington, J. P. ChEMBL: A Large-Scale Bioactivity Database for Drug Discovery. *Nucleic Acids Res.* **2012**, *40* (Database issue), D1100–D1107.
(92) Groom, C. R.; Bruno, I. J.; Lightfoot, M. P.; Ward, S. C. The Cambridge Structural Database. *Acta Crystallogr B Struct Sci Cryst Eng Mater* **2016**, *72* (Pt 2), 171–179.
(93) Wishart, D. S.; Knox, C.; Guo, A. C.; Shrivastava, S.; Hassanali, M.; Stothard, P.; Chang, Z.; Woolsey, J. DrugBank: A Comprehensive Resource for in Silico Drug Discovery and Exploration. *Nucleic Acids Res.* **2006**, *34* (Database issue), D668–D672.
(94) Otsuka, S.; Kuwajima, I.; Hosoya, J.; Xu, Y.; Yamazaki, M. PoLyInfo: Polymer Database for Polymeric Materials Design. In *2011 International Conference on Emerging Intelligent Data and Web Technologies*; ieeexplore.ieee.org, 2011; pp 22–29.
(95) A Community Resource for Innovation in Polymer Technology https://cript.mit.edu/.

**Supporting Information**

# A graph representation of molecular ensembles for polymer property prediction

Matteo Aldeghi [1], Connor W. Coley [1,2,*]

[1] Department of Chemical Engineering, Massachusetts Institute of Technology, Cambridge, MA 02139, USA

[2] Department of Electrical Engineering and Computer Science, Massachusetts Institute of Technology, Cambridge, MA 02139, USA

[*] Email: ccoley@mit.edu

This PDF includes:
- Extended Methods (pp. S2–S5)
- Figures S1–S3 (pp. S6–S8)
- Table S1–S2 (pp. S9–S10)

Other supporting materials for this manuscript include the following:
- Data S1: a dataset of computed electron affinity and ionization potential values for 42,966 copolymers (CSV). The files includes SMILES strings for monomers A and B ("monoA" and "monoB" headers), their relative mole fractions ("fracA" and "fracB"), the copolymer chain architecture ("chain_arch"), the raw electron affinity and ionization potential values computed with xTB ("EA (ev)" and "IP (eV)"), and their calibrated values with respect to the standard hydrogen potential ("EA vs SHE (eV)" and "IP vs SHE (eV)").
- An implementation of the wD-MPNN at: https://github.com/coleygroup/polymer-chemprop.
- Data and scripts needed to reproduce the results in this paper at: https://github.com/coleygroup/polymer-chemprop-data.

# Extended Methods

## *Copolymer dataset*

For each copolymer considered, we generated up to 32 sequences and 8 conformers per sequence. Sequences were generated at random using custom *RDKit*[1] (v2021.03) scripts, while maintaining the specified chain architecture and monomer ratio. When more than 32 sequences were possible, only 32 at random were kept. Conformers were generated using the ETKDG approach[2] as implemented in *RDKit*, while automatically falling back to ETDG[2] in rare failure cases. All conformers were then minimized with the MMFF94 force field[3–7], also via *RDKit*.

All the octamer structures generated as described above were further minimized with GFN2-xTB[8] using the *xtb* program[9], with energy tolerance of 5 × 10⁻⁵ Eh and gradient tolerance of 4 × 10⁻³ Eh/α. Vertical IP and EA properties were then computed with IPEA-xTB[10], a specially reparameterized version of GFN1-xTB[11]. All xTB calculations used the analytical linearized Poisson-Boltzmann model for implicit solvation[12,13] with the default parameters for benzene in the *xtb* code. The IP and EA values were Boltzmann averaged across the 8 conformers at 298.15 K, and then averaged across all sequences associated with a specific copolymer. Finally, IP and EA were adjusted to reflect their relative values with respect to the standard hydrogen electrode potential (4.44 V), and then corrected according to the DFT calibration in a low dielectric environment performed by Wilbraham *et al.*[14].

## *Architecture of the wD-MPNN*

Consider a molecular graph $G = \{V, E\}$ with vertices $V$, edges $E$, node features $x_v$, and edge features $e_{vu}$, where $v \in V$ and $vu \in E$. Node features are initialized as a one-hot encoding of the atomic number, degree, formal charge, chirality, number of hydrogens, hybridization, and aromaticity of the atom, and scaled atomic mass of each atom. Bond features are initialized as a one-hot encoding of the bond type (singe, double, triple, or aromatic), whether the bond is conjugated, in a ring, and contains stereochemical information. The hidden features of directed edges $h_{vu}^0$ are built by concatenating the features $x_v$ and $e_{vu}$ (i.e., those of each atom and its outgoing edge) and passing the resulting vector through a single neural network layer

$$h_{vu}^0 = \tau(W_i \text{cat}(x_v, e_{vu})),$$

where $\tau$ is the an activation function (here chosen to be ReLU), and $W_i \in \mathbb{R}^{h \times h_i}$ is a learned matrix with $h$ being a user-defined hidden size (here chosen to be 300) and $h_i$ the size of $\text{cat}(x_v, e_{vu})$. These hidden features are then updated via repeated message passing operations. Here we will use a total of $T = 3$ message passing steps. At very step $t \in \mathbb{Z}_{<T}$, updated features are obtained as

$$h_{vu}^{t+1} = \tau\left(h_{vu}^0 + W_h \sum_{k \in \{N(v) \setminus u\}} w_{kv} h_{kv}^t\right),$$

where $W_h \in \mathbb{R}^{h \times h}$ is a learned matrix, $N(v) \setminus u$ are the neighbors of $v$ aside from $u$, and $w_{kv}$ are user-provided directed edge weights according to the probability of $v$ being a neighbor of $u$ in the polymer graph (Figure 1). The addition of $h_{vu}^0$ at every step acts as a skip connection. For a standard molecule, $w_{vu} = 1 \ \forall \ v, u \in V$, but for a polymer these weights allow us to distinguish between different monomer sequences, as they weigh the messages $m_{kv}^{t+1} = \sum_{k \in \{N(v) \backslash u\}} w_{kv} h_{kv}^t$ according to the relevance of each incoming edge.

After $T$ message passing iterations, atoms representations $h_v$ are obtained by summing over all incoming edge representation $h_k^T v$, concatenating them with the original atomic features $x_v$, and passing the resulting vector through a single neural network layer,

$$h_v = \tau \left( W_o \ \text{cat} \left( x_v, \sum_{k \in N(v)} w_{kv} h_{kv}^T \right) \right),$$

where $W_o \in \mathbb{R}^{h \times h_o}$ is a learned matrix, $h_o$ is the size of the concatenated vector, and $w_{kv}$ are the weights of the incoming edges. In such a way, the output atom features depend on the probability of each incoming edge being present in the molecule ensemble. Finally, a molecule feature vector is obtained as an average (or sum) of all atom feature vectors,

$$h = \frac{1}{|V|} \sum_{v \in G} w_v h_v,$$

where $w_v$ are weights reflecting the stoichiometric ratio between different monomers. The learned representation $h$ is then used as the input of a feed-forward neural network to predict the molecular or polymer properties of interest. In our tests, we use a shallow network with two layers, where the first is the input layer with hidden dimension $h$, and the second is the output layer with dimension matching the number of labels being predicted.

### *Baseline representations and models*

The predictive performance of wD-MPNN developed was compared to that of established ML approaches. These included the base D-MPNN model (standard Chemprop[15]) and random forest (RF) models trained on fingerprint representations. All models were trained as multi-task models aimed at predicting both EA and IP.

**Baselines for the EA and IP dataset**

The D-MPNN baseline used a model with the same architecture described in Figure 2, but without information on monomer stoichiometry or the partial bonds connecting repeating units. The input for this model was a disconnected graph of the separate monomeric units. In Table 1, this model is referred to as "Monomer", because it only captures the chemical structure of the monomeric units. The extension of this model in which weighted edges are used is referred to as "Monomer + Chain architecture". The extension of this model in which weighted atomic embeddings are used is referred to as "Monomer + Stoichiometry". The proposed wD-MPNN includes both these features and is thus referred to in Table 1 as "Monomer + Chain architecture + Stoichiometry".

Extended-Connectivity Fingerprints (ECFP)[16] were used as input representations for RF models, as implemented in *scikit-learn*[17] (v1.0). These were obtained with *RDKit*[1] using 2048 bits and radius 2.

We tested both binary and count fingerprints, constructed from the monomeric units alone, as well as from an ensemble of oligomeric sequences sampled uniformly at random. In the first case, the monomeric units were converted to a single *RDKit* molecule object and the fingerprints computed. This is the fingerprint equivalent of the representation provided to the baseline D-MPNN model. In the second case, we sampled up to 32 octameric sequences, while satisfying the stoichiometry and chain architecture of the polymer (Figure 3), computed fingerprints for all resulting oligomers, and used their average as input for the RF model. This fingerprint representation of polymers, which is proposed in this study, attempts to capture the ensemble nature of the polymeric material. Despite the sampling procedure, binary fingerprints are not able to capture information on stoichiometry, while count fingerprints can. As shown in the Results section, this latter approach was found to be the most competitive among all baselines considered. Patel *et al*.[18] and Kuenneth *et al*.[19] proposed the use of a weighted sum of the binary fingerprints of each monomer to capture the stoichiometry of different monomer units in a copolymer. This approach is somewhat similar to ours based on count fingerprints and multiple sampled sequences, however, ours captures information on chain architecture in addition to stoichiometry.

All models were evaluated on the same cross-validation splits, which included train, validation, and test sets. Both random and monomer splits were evaluated, as discussed in the Results. No hyperparameter optimization was performed for the D-MPNN and wD-MPNN models and the validation set was used only for early stopping. The RF models were trained in two ways that differed in how the validation set was used. In the first approach, RF models with default *scikit-learn* parameters (100 trees) were trained on the concatenation of the training and validation sets. In the second approach, the RF hyperparameters were tuned against the validation set with 20 iterations of Bayesian optimization using *Hyperopt*[20], in which the number of trees (100 to 500), the maximum tree depth (5 to 20), the minimum number of samples required to split a node (2 to 6), and the minimum number of samples required to be at a leaf node (1 to 5), were optimized. Because the first approach was found to provide higher test-set performance overall (Table S2), we discuss only those results in the main text. While this selection does slightly inflate the estimated performance of the RF baselines, we found these to still be significantly inferior to those obtained with the wD-MPNN model.

**Baselines for the diblock copolymer phases dataset**
The D-MPNN baselines used in these experiments are the same as those used for the EA and IP prediction tasks, as described above. However, in the wD-MPNN we also introduced information on polymer size by scaling the molecular embeddings $h$ by $1 + log(N)$, where $N$ is the degree of polymerization. We thus considered an additional baseline model, referred to as "Monomers + Size" in Figure 5, in which the base D-MPNN also scales its molecular vector representations by chain length. The degree of polymerization for the diblock copolymer, as well as average chain lengths for each block, were obtained from the number-average molar mass of the copolymer, the volume fractions of each block, and the bulk densities of each block. Molar mass and volume fractions are reported for all dataset entries in the Block Copolymer Phase Behavior Database[21], while bulk densities were taken from Table S1 of Arora *et al.*[22] for all entries in which they were missing.

RF models (Figure S3) as implemented in *scikit-learn*[17] (v1.0) were used with count-vector Extended-Connectivity Fingerprints[16] with radius 2 and 2048 bits using *RDKit*[1]. Fingerprint representations in which only the chemical structure of the monomers was considered are referred to as "Monomers" in Figure S3, consistently with naming used for the D-MPNN tests. In this case, the count fingerprints of the homopolymers or copolymers constituting each block were obtained separately and then summed. The effect of chain architecture was captured by building 32 cyclic dodecameric sequences by sampling from the available monomers uniformly at random, for each of the two polymer blocks separately. The average of the count fingerprints across the 32 sequences was then taken as the representation of each block. The averaged fingerprints from each of the two blocks was then summed to obtain the final diblock copolymer representation. This input representation is referred to as "Monomers + Chain architecture". The effect of block stoichiometry was captured by scaling the count fingerprints of each block by the block's mole ratio. This input representation is referred to as "Monomers + Stoichiometry". The effect of copolymer size was considered by scaling the fingerprint representation of the diblock copolymer (i.e., after summing the fingerprints of each block) by $1 + log(N)$, as it was done for the D-MPNN representations. This input representation is referred to as "Monomers + Size" in Figure S3. The richest fingerprint-based representation considered was called "Monomers + Chain architecture + Size + Stoichiometry", and used all of the above transformations of the base fingerprints representation. That is, for each block, count fingerprints averaged across 32 sampled sequences were obtained. These were then scaled by the mole fractions of the blocks, then summed, and then scaled proportionally to chain length.

In addition to the above, also pure descriptor-based representations that disregard the chemistry of the copolymer blocks were tested in conjunction with RF models. These are tagged as "no chem" in Figure S3. In this case, the input for the RF was the mole fraction of block A ("Stoichiometry") only, the average chain length of the copolymer ("Size") only, or both of these ("Size + Stoichiometry").

All models for this task were evaluated on the same stratified 10-fold cross-validation splits, in which the fraction of positive labels for each class were approximately preserved across folds. As for the EA/IP regression task, each fold included train, validation, and test set. For the D-MPNN models, the validation set was used only for early stopping, while RF models were trained on both training and validation sets together. RF classification models used 100 trees (`n_estimators=100`), the Gini criterion to evaluate node split quality (`criterion="gini"`), balanced class weights (`class_weight="balanced"`), and nodes were expanded until all leaves were pure (`max_depth=None`).

**Figure S1 | Performance of the wD-MPNN and baseline models for the prediction of electron affinity (EA) and ionization potential (IP) on a monomer 9-fold cross-validation split.** Each parity plot shows the computed DFT values against the ones predicted by the ML models. The parity line is shown as a black dashed line. The scatter/density plots display the predictions of each model for all folds of the 9-fold cross validation. Each fold contains a monomer A (Figure 3) that is absent from all other folds. The average coefficient of determination ($R^2$) and root-mean-square error (RMSE, in eV) across all folds are shown. (a) Performance of the baseline D-MPNN model, which used a graph representation of the monomeric units as input. (b) Performance of the wD-MPNN model, which is augmented with information about chain architecture and monomer stoichiometry. (c) Performance of a random forest (RF) model that used a binary fingerprint (FP) representation of the monomeric units as input. (d) Performance of a RF model that used a binary fingerprint representation of the polymer as input, which was obtained as the average fingerprint of a set of oligomeric sequences sampled uniformly at random, while satisfying the correct stoichiometry and chain architecture of the specified polymer. (e) Performance of a RF model that used a count fingerprint representation of the monomeric units as input. (f) Performance of a RF model that used a count fingerprint representation of the polymer as input.

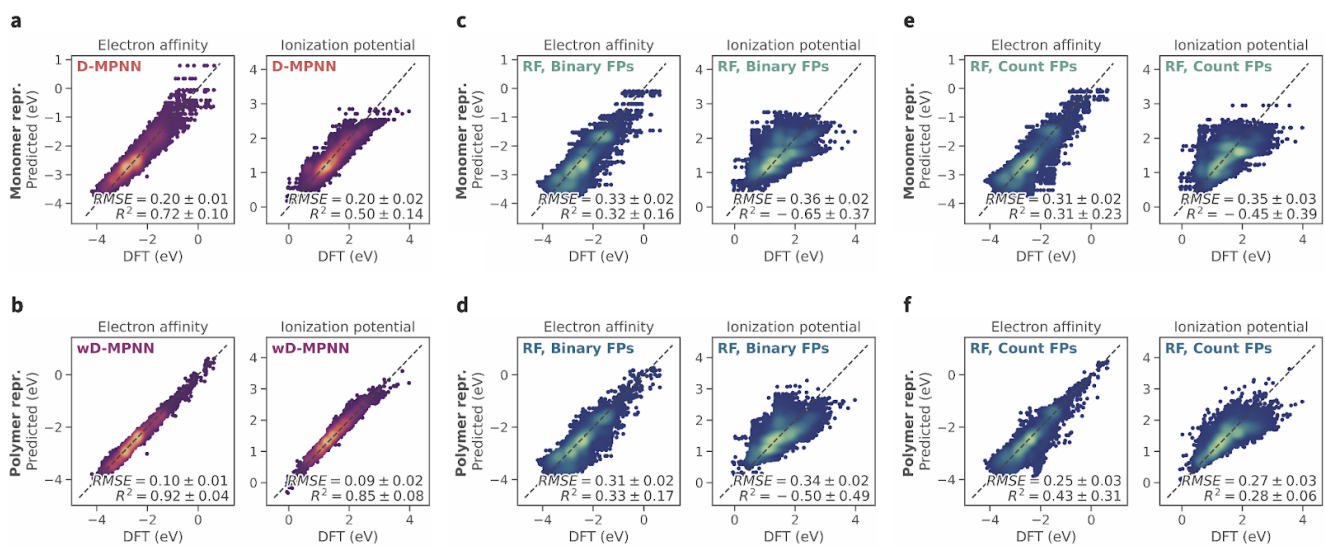

**Figure S2 | Performance of the wD-MPNN model and baselines at different levels of data availability for training.** Among the RF baselines, only results for the approach that used a count fingerprint representation of the polymer as input (i.e., the most competitive RF baseline found) are shown. The plots show the test set performance of these three models for multiple random splits, in which the size of the training set was increased from 0.1% (43) to 80% (34,373) of the whole dataset. The size of the validation set was kept constant at 10% (4,296). All remaining data was used as part of the test set, which thus had size between 89.9% (38,626) to 10% (4,296). 10 random splits were used. The markers in the plots show the mean test set performance across these repeated splits. Standard errors of the mean are not shown because they are small and would not be visible. The largest $R^2$ error is 0.03, and the largest RMSE error is 0.01 eV. Overall, these results show a cross-over point, at a training set size of 2% (859), in terms of performance between the wD-MPNN and RF approaches. While RF has better performance in the lower-data regime, the wD-MPNN overtakes RF when more data is available. Similarly, the performance of D-MPNN and wMPNN starts diverging after 1% (430) of the data is available for training.

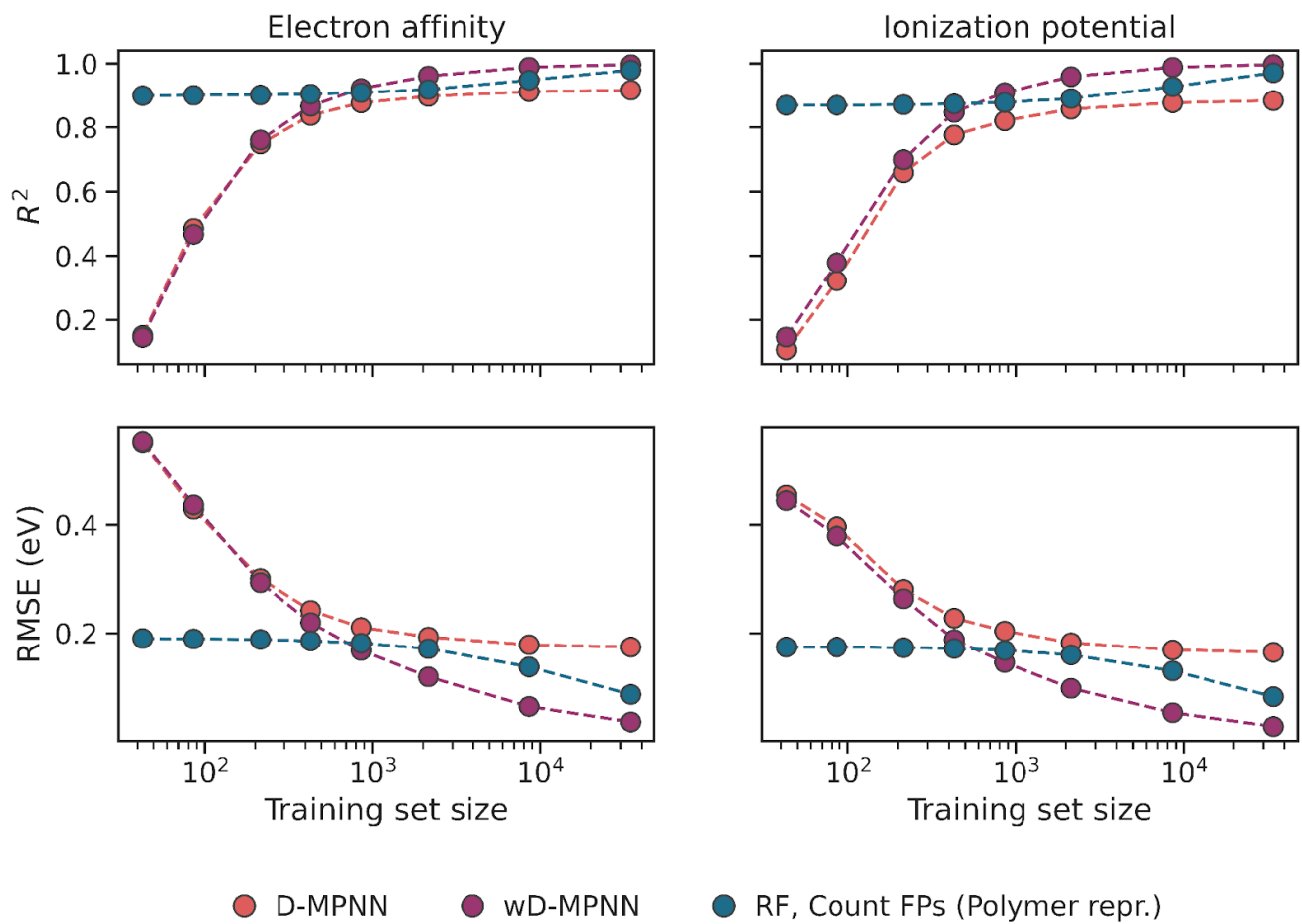

**Figure S3 | Performance of multiple models and input representations on the classification of diblock copolymer phases.** Performance is measured as the average area under the precision-recall curve (PRC) across the five labels (phases) to be predicted. Each marker reflects the average PRC value, for one fold of a 10-fold cross validation, across the five classification labels. The boxes show the first, second, and third quartiles of the data, and the whiskers extend to up to 1.5 times the interquartile range. Models are grouped into (i) D-MPNN, in which the one labeled as "Monomers + Chain arch. + Size + Stoichiometry" corresponds to the full wD-MPNN, (ii) RF (no chem.), in which only one or two descriptors are used as input, without considering the chemical structure of the constituting monomers, and (iii) RF, in which the chemical structure of the monomers, as well as different aspects of the polymer, are considered. A detailed explanation of each model and input is provided in the SI Extended Methods.

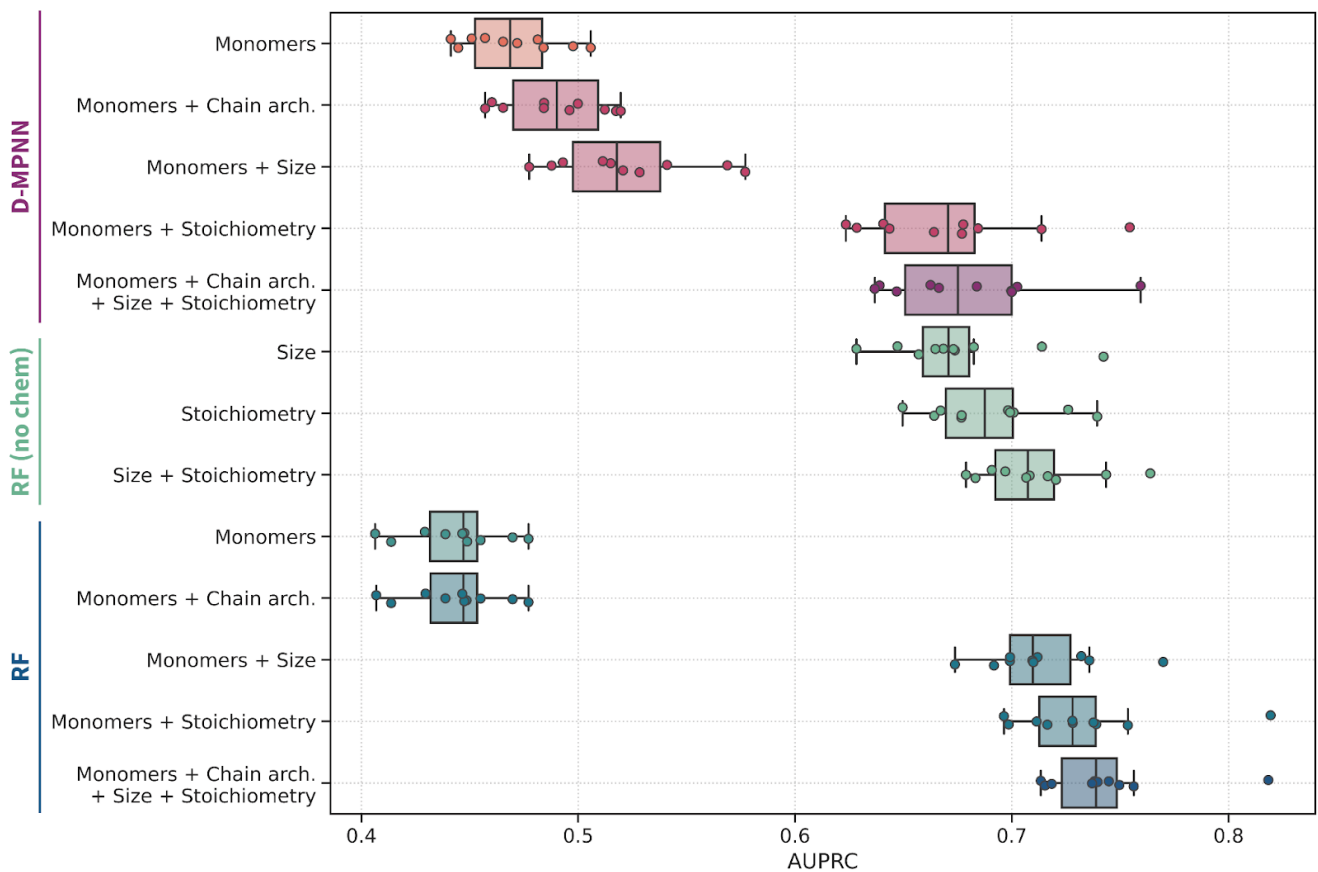

**Table S1 | Performance of wD-MPNN and ablated architectures for the prediction of copolymer EA and IP values on the original and artificially inflated datasets.** The average $R^2$ and RMSE values obtained from 10-fold cross validation based on random splits, for the prediction of IP and EA, are shown. The standard error of the mean is reported in parenthesis and it applies to the least significant digits (e.g., 0.917(1) is equivalent to 0.917 ± 0.001). Under the header "Representation", "monomers" indicates the model was provided with the graph structure of separate monomer units (i.e., the baseline D-MPNN); "chain architecture" indicates the model was provided with information on how the monomer units may connect to one another to form an ensemble of possible sequences, via the definition of edge weights, used as shown in Figure 2b and 2c; "stoichiometry" indicates the model was provided with information on monomer stoichiometry, which was used to weigh learnt node representations as shown in Figure 2d. The model using all this information altogether corresponds to the wD-MPNN. Under the header "Dataset", results for three different datasets are reported. The original dataset used EA and IP values as they were computed. In the dataset with inflated chain architecture importance, the standard deviation of EA and IP values was increased by a factor of 5 for any combination of monomer pair and stoichiometry. In the dataset with inflated chain architecture importance, the standard deviation of EA and IP values was increased by a factor of 5 for any combination of monomer pairs and chain architecture.

| | | Dataset | | | | | |
|---|---|---|---|---|---|---|---|
| | | Original dataset | | Inflated chain architecture importance | | Inflated stoichiometry importance | |
| **Representation** | **Prop.** | $R^2$ | RMSE (eV) | $R^2$ | RMSE (eV) | $R^2$ | RMSE (eV) |
| Monomers | EA | 0.917(1) | 0.173(1) | 0.735(3) | 0.343(2) | 0.355(4) | 0.760(3) |
| | IP | 0.883(1) | 0.165(1) | 0.647(3) | 0.333(2) | 0.265(4) | 0.737(2) |
| Monomers + Chain arch. | EA | 0.929(1) | 0.159(1) | 0.890(1) | 0.221(1) | 0.356(4) | 0.759(3) |
| | IP | 0.898(1) | 0.154(1) | 0.861(1) | 0.208(1) | 0.266(3) | 0.736(2) |
| Monomers + Stoichiometry | EA | 0.987(0) | 0.069(0) | 0.788(2) | 0.307(2) | 0.971(0) | 0.161(1) |
| | IP | 0.982(0) | 0.065(1) | 0.714(2) | 0.299(2) | 0.972(0) | 0.144(1) |
| Monomers + Chain arch. + Stoichiometry | EA | 0.997(0) | 0.035(0) | 0.978(1) | 0.098(1) | 0.985(0) | 0.115(1) |
| | IP | 0.997(0) | 0.027(0) | 0.981(0) | 0.077(1) | 0.988(0) | 0.094(1) |

**Table S2 | Performance of RF models with and without hyperparameter optimization.** Average cross-validated RMSE values are shown. The standard error of the mean is reported in parenthesis and it applies to the least significant digits (e.g., 0.327(25) is equivalent to 0.327 ± 0.025). The use of hyperparameter optimization is indicated by the "w/ opt." label. In this case 20 optimization iterations with *Hyperopt* were performed. The use of a fixed set of hyperparameters is indicated by the "w/o opt." label. For this task, and a search limited to 20 iterations, there did not seem to be a performance gain with hyperparameter tuning. In fact, default hyperparameters returned slightly better performance on average.

|  |  | Cross-validation split | | | |
|---|---|---|---|---|---|
|  |  | Random split | | Monomer split | |
| **Approach** |  | *EA (eV)* | *IP (eV)* | *EA (eV)* | *IP (eV)* |
| RF, Binary FPs Monomer Repr. | w/o opt. | 0.188(1) | 0.179(1) | 0.327(25) | 0.362(18) |
|  | w/ opt. | 0.190(1) | 0.181(1) | 0.354(32) | 0.399(36) |
| RF, Count FPs Monomer Repr. | w/o opt. | 0.188(1) | 0.179(1) | 0.306(24) | 0.353(34) |
|  | w/ opt. | 0.187(1) | 0.178(1) | 0.323(26) | 0.378(47) |
| RF, Binary FPs Polymer Repr. | w/o opt. | 0.151(1) | 0.145(1) | 0.311(19) | 0.340(23) |
|  | w/ opt. | 0.163(1) | 0.154(1) | 0.355(15) | 0.388(41) |
| RF, Count FPs Polymer Repr. | w/o opt. | 0.086(1) | 0.083(1) | 0.251(25) | 0.273(32) |
|  | w/ opt. | 0.107(1) | 0.104(1) | 0.273(30) | 0.297(50) |

# References

(1) Landrum, G. A. *RDKit: Open-Source Cheminformatics*.
(2) Riniker, S.; Landrum, G. A. Better Informed Distance Geometry: Using What We Know To Improve Conformation Generation. *J. Chem. Inf. Model.* **2015**, *55* (12), 2562–2574.
(3) Halgren, T. A. Merck Molecular Force Field. I. Basis, Form, Scope, Parameterization, and Performance of MMFF94. *Journal of Computational Chemistry* **1996**, *17* (5-6), 490–519.
(4) Halgren, T. A. Merck Molecular Force Field. II. MMFF94 van Der Waals and Electrostatic Parameters for Intermolecular Interactions. *J. Comput. Chem.* **1996**, *17* (5-6), 520–552.
(5) Halgren, T. A. Merck Molecular Force Field. III. Molecular Geometries and Vibrational Frequencies for MMFF94. *J. Comput. Chem.* **1996**, *17* (5-6), 553–586.
(6) Thomas A. Halgren, R. B. N. Merck Molecular Force Field. IV. Conformational Energies and Geometries for MMFF94. *J. Comput. Chem.* **1996**, *17* (5-6), 587–615.
(7) Halgren, T. A. Merck Molecular Force Field. V. Extension of MMFF94 Using Experimental Data, Additional Computational Data, and Empirical Rules. *J. Comput. Chem.* **1996**, *17* (5-6), 616–641.
(8) Bannwarth, C.; Ehlert, S.; Grimme, S. GFN2-xTB—An Accurate and Broadly Parametrized Self-Consistent Tight-Binding Quantum Chemical Method with Multipole Electrostatics and Density-Dependent Dispersion Contributions. *J. Chem. Theory Comput.* **2019**, *15* (3), 1652–1671.
(9) P. Atkinson, C. Bannwarth, F. Bohle, G. Brandenburg, E. Caldeweyher, M. Checinski, S. Dohm, S. Ehlert, S. Ehrlich, I. Gerasimov, S. Grimme, C. Hölzer, A. Katbashev, J. Koopman, C. Lavinge, S. Lehtola, F. März, M. Müller, F. Musil, H. Neugebauer, J. Pisarek, C. Plett, P. Pracht, F. Pultar, J. Seibert, P. Shushkov, S. Spicher, M. Stahn, M. Steiner, T. Strunk, J. Stückrath, T. Rose, J. Unsleber. *Xtb*.
(10) Ásgeirsson, V.; Bauer, C. A.; Grimme, S. Quantum Chemical Calculation of Electron Ionization Mass Spectra for General Organic and Inorganic Molecules. *Chem. Sci.* **2017**, *8* (7), 4879–4895.
(11) Grimme, S.; Bannwarth, C.; Shushkov, P. A Robust and Accurate Tight-Binding Quantum Chemical Method for Structures, Vibrational Frequencies, and Noncovalent Interactions of Large Molecular Systems Parametrized for All Spd-Block Elements (Z = 1–86). *J. Chem. Theory Comput.* **2017**, *13* (5), 1989–2009.
(12) Fogolari, F.; Brigo, A.; Molinari, H. The Poisson-Boltzmann Equation for Biomolecular Electrostatics: A Tool for Structural Biology. *J. Mol. Recognit.* **2002**, *15* (6), 377–392.
(13) Tuinier, R. Approximate Solutions to the Poisson–Boltzmann Equation in Spherical and Cylindrical Geometry. *J. Colloid Interface Sci.* **2003**, *258* (1), 45–49.
(14) Wilbraham, L.; Berardo, E.; Turcani, L.; Jelfs, K. E.; Zwijnenburg, M. A. High-Throughput Screening Approach for the Optoelectronic Properties of Conjugated Polymers. *J. Chem. Inf. Model.* **2018**, *58* (12), 2450–2459.
(15) Yang, K.; Swanson, K.; Jin, W.; Coley, C.; Eiden, P.; Gao, H.; Guzman-Perez, A.; Hopper, T.; Kelley, B.; Mathea, M.; Palmer, A.; Settels, V.; Jaakkola, T.; Jensen, K.; Barzilay, R. Analyzing Learned Molecular Representations for Property Prediction. *J. Chem. Inf. Model.* **2019**, *59* (8), 3370–3388.
(16) Rogers, D.; Hahn, M. Extended-Connectivity Fingerprints. *J. Chem. Inf. Model.* **2010**, *50* (5), 742–754.
(17) Pedregosa, F.; Varoquaux, G.; Gramfort, A.; Michel, V.; Thirion, B.; Grisel, O.; Blondel, M.; Prettenhofer, P.; Weiss, R.; Dubourg, V.; Vanderplas, J.; Passos, A.; Cournapeau, D.; Brucher, M.; Perrot, M.; Duchesnay, É. Scikit-Learn: Machine Learning in Python. *J. Mach. Learn. Res.* **2011**, *12* (85), 2825–2830.
(18) Patel, R. A.; Borca, C. H.; Webb, M. A. Featurization Strategies for Polymer Sequence or Composition Design by Machine Learning. *Mol. Syst. Des. Eng.* **2022**, -.
(19) Kuenneth, C.; Schertzer, W.; Ramprasad, R. Copolymer Informatics with Multitask Deep Neural Networks. *Macromolecules* **2021**, *54* (13), 5957–5961.
(20) Bergstra, J.; Yamins, D.; Cox, D. Making a Science of Model Search: Hyperparameter Optimization in Hundreds of Dimensions for Vision Architectures. In *Proceedings of the 30th*

*International Conference on Machine Learning*; Dasgupta, S., McAllester, D., Eds.; Proceedings of Machine Learning Research; PMLR: Atlanta, Georgia, USA, 17--19 Jun 2013; Vol. 28, pp 115–123.
(21) Rebello, N. J.; Arora, A.; Mochigase, H.; Lin, T.-S.; Audus, D. J.; Olsen, B. D. Block Copolymer Phase Behavior Database.
(22) Arora, A.; Lin, T.-S.; Rebello, N. J.; Av-Ron, S. H. M.; Mochigase, H.; Olsen, B. D. Random Forest Predictor for Diblock Copolymer Phase Behavior. *ACS Macro Lett.* **2021**, *10* (11), 1339–1345.